\documentclass{article}
\usepackage{amsmath,amsfonts, amssymb, mathtools, bbm, nicefrac}

\newcommand{\E}{\mathbb{E}}

\DeclareMathAlphabet{\mathsfit}{\encodingdefault}{\sfdefault}{m}{sl}
\SetMathAlphabet{\mathsfit}{bold}{\encodingdefault}{\sfdefault}{bx}{n}

\def\sR{{\mathbb{R}}}

\usepackage[dvipsnames]{xcolor} 
\definecolor{darkblue}{rgb}{0.0, 0.0, 0.5}

\usepackage{hyperref}

\usepackage{graphicx, wrapfig}
\usepackage{subcaption}
\usepackage{multirow,booktabs}

\usepackage{microtype}
\usepackage{ulem}
\usepackage{soul}
\usepackage[group-separator={,}]{siunitx}
\usepackage{arydshln}

\newcommand{\tit}[1]{\textit{#1}}
\newcommand{\tbf}[1]{\textbf{#1}}

\usepackage{cancel}

\newcommand\blfootnote[1]{%
	\begingroup
	\renewcommand\thefootnote{}\footnote{#1}%
	\addtocounter{footnote}{-1}%
	\endgroup
}

\usepackage[accepted]{icml2021}

\begin{document}
\icmltitlerunning{Improving Adversarial Robustness of CNNs via CIFS}
\twocolumn[
\icmltitle{CIFS: Improving Adversarial Robustness of CNNs via Channel-wise Importance-based Feature Selection}
\icmlsetsymbol{equal}{*}
\begin{icmlauthorlist}
	\icmlauthor{Hanshu Yan}{nus-ece}
	\icmlauthor{Jingfeng Zhang}{riken-aip}
	\icmlauthor{Gang Niu}{riken-aip}
	\icmlauthor{Jiashi Feng}{nus-ece}
	\icmlauthor{Vincent Y. F. Tan}{nus-ece,nus-math}
	\icmlauthor{Masashi Sugiyama}{riken-aip,u-tokyo}
\end{icmlauthorlist}
\icmlaffiliation{nus-ece}{Department of Electrical and Computer Engineering, National University of Singapore, Singapore;}
\icmlaffiliation{nus-math}{Department of Mathematics, National University of Singapore, Singapore;}
\icmlaffiliation{riken-aip}{RIKEN Center for Advanced Intelligence Project (AIP), Tokyo, Japan;}
\icmlaffiliation{u-tokyo}{Graduate School of Frontier Sciences, The University of Tokyo, Tokyo, Japan}
\icmlcorrespondingauthor{Jingfeng Zhang}{jingfeng.Zhang@riken.jp}
\icmlkeywords{Machine Learning, ICML}
\vskip 0.3in
]


\printAffiliationsAndNotice{}  

\begin{abstract}
We investigate the adversarial robustness of CNNs from the perspective of channel-wise activations.  By comparing normally trained and adversarially trained models, we observe that adversarial training (AT) robustifies CNNs by aligning the channel-wise activations of adversarial data with those of their natural counterparts. However, the channels that are \textit{negatively-relevant} (NR) to predictions are still over-activated when processing adversarial data. Besides, we also observe that AT does not result in similar robustness for all classes. For the robust classes, channels with larger activation magnitudes are usually more \textit{positively-relevant} (PR) to predictions, but this alignment does not hold for the non-robust classes. Given these observations, we hypothesize that suppressing NR channels and aligning PR ones with their relevances further enhances the robustness of CNNs under AT. To examine this hypothesis, we introduce a novel mechanism, \textit{i.e.}, \underline{C}hannel-wise \underline{I}mportance-based \underline{F}eature \underline{S}election (CIFS). The CIFS manipulates channels' activations of certain layers by generating non-negative multipliers to these channels based on their relevances to predictions. Extensive experiments on benchmark datasets including CIFAR10 and SVHN clearly verify the hypothesis and CIFS's effectiveness of robustifying CNNs.
\end{abstract}

\section{Introduction}\label{introduction}

Convolutional neural networks (CNNs) have achieved tremendous successes in real-world applications, such as  autonomous vehicles \cite{grigorescu2020survey, hu2020panda} and computer-aided medical diagnoses \cite{trebeschi2017deep, shu2020enhancing}. However, CNNs have be shown vulnerable to well-crafted (and even minute) adversarial perturbations to inputs \cite{szegedy_intriguing_2014, goodfellow_explaining_2015, ilyas_adversarial_2019}. This has become hazardous in high-stakes applications such as medical diagnoses and autonomous vehicles.\blfootnote{Code: \url{https://github.com/HanshuYAN/CIFS}}

Recently, many empirical defense methods have been proposed to secure CNNs against these adversarial perturbations, such as adversarial training (AT) \cite{madry_towards_2019}, input/feature denoising \cite{xie_feature_2019, du_rain_2020} and defensive distillation \cite{papernot_distillation_2016}. AT~\cite{madry_towards_2019}, which generates adversarial data on the fly for training CNNs, has emerged as one of the most successful methods. AT effectively robustifies CNNs but leads to a clear drop in the accuracies for natural data~\cite{tsipras_robustness_2018} and suffers from the problem of overfitting to adversarial data used for training \cite{rice_overfitting_2020,zhang_geometry-aware_2020,chen2021robust}. To ameliorate these problems, researchers have proposed variants of AT, including TRADES \cite{zhang_theoretically_2019} and Friendly-Adversarial-Training (FAT) \cite{zhang_attacks_2020}. 
To further robustify CNNs under AT, many works attempt to propose novel defense mechanism to mitigate the effects of adversarial data on features \cite{xie_feature_2019, du_rain_2020, xu_interpreting_2019}. For example, \citet{xie_feature_2019} found that adversarial data result in abnormal activations in the feature maps and performed feature denoising to remove the adversarial effects. Most of these works improved robustness by identifying and suppressing abnormalities \tit{at certain positions} across channels~(commonly referred to as feature maps in CNNs), whereas the other direction, namely, the connection between robustness and irregular activations of certain \tit{entire channels}, has received scant attention.

Since channels of CNNs' deeper layers are capable of extracting semantic characteristic features \cite{zeiler_visualizing_2014}, the process of making predictions usually relies heavily on aggregating information from various channels \cite{bach2015pixel}. As such, anomalous activations of certain channels may result in incorrect predictions. Thus, it is imperative to explore 
which channels are \tit{entirely} irregularly activated by adversarial data and which channels' activations benefit or degrade robustness.
By utilizing this connection, we will be able to further enhance the robustness of CNNs via suppressing or promoting certain vulnerable or reliable channels respectively. 

In this work, we attempt to build such a connection by comparing the channel-wise activations of non-robust (normally trained) and robustified (adversarially trained) CNNs. The \tit{channel-wise activations} are defined as the average activation magnitudes of all features within channels \cite{bai_improving_2021}. To identify what types of channels appear to be abnormal under attacks, we regard \tit{channels' relevances} to prediction results (formally defined in Equation (\ref{eq:relevance-assessment}) as $g^l$) as the gradients of the corresponding logits \tit{w.r.t} channel-wise activations. The channels, whose relevances to prediction results are positive or negative ($g^l_{[i]}>0$ or $g^l_{[i]}\leq 0$), are called \textit{positively-relevant} (PR) or \textit{negatively-relevant} (NR) channels.

On the one hand, we observe that, AT robustifies CNNs by aligning adversarial data's channel-wise activations with those of natural data. However, we find that the NR channels of adversarially trained CNNs are still over-activated by adversarial data (see Figure \ref{fig: w-vs-act_adv-auto}). Thus, we wonder: \textit{If we suppress NR channels during AT to facilitate the alignment of channel’s activations, will it benefit CNNs' robustness?} 
On the other hand, we find that adversarially trained classification models do not enjoy similar robustness across all the classes (see Figure \ref{fig: w-vs-act_adv-auto} and \ref{fig: w-vs-act_adv-cat}). For classes with relatively good robustness, channels' activations usually align well with their relevances, i.e., channels with larger activations are more PR to labels. Given this phenomenon, a natural question arises: \textit{If we align channels' activations with their relevances during AT, will it improve the robustness of CNNs?} Regarding these two questions, we propose a unified hypothesis on robustness enhancement, denoted as $\mathcal H$: \ul{Suppressing NR channels and aligning channels' activations with their relevances to prediction results benefit the robustness of CNNs}.

To examine this hypothesis, we propose a novel mechanism, called \underline{C}hannel-wise \underline{I}mportance-based \underline{F}eature \underline{S}election (CIFS), which adjusts channels' activations with an \tit{importance mask} generated from channels' relevances. For a certain layer, CIFS first takes as input the representation of a data point at this layer and makes a \textit{raw prediction} for the data point by a \textit{probe network}. The probe serves as the surrogate for the subsequent classifier (the composition of subsequent layers) in the backbone and is jointly trained with the backbone under supervision of true labels. Then, CIFS computes the gradients of the sum of the top-$k$ logits \textit{w.r.t.} the channels' activations. We can obtain the \tit{relevance} of each channel to the top-$k$ prediction results by accumulating the gradients within the channel. Finally, CIFS generates a mask of importance scores for each channel by mapping channels' relevances monotonically to non-negative values. 
Through extensive experiments, we answer the two questions in the affirmative and confirm hypothesis $\mathcal H$. Indeed, our results show that CIFS clearly enhances the adversarial robustness of CNNs. 

We comprehensively evaluate the robustness of CIFS-modified CNNs on benchmark datasets against various attacks. On the CIFAR10 dataset, CIFS improves the robustness of the ResNet-18 by $4$ percentage points against the PGD-100 attack. We also observe that CIFS ameliorates the overfitting during AT. In particular, the robustness at the last epoch is close to that at the best epoch. Finally, we conduct an ablation study to further understand how various elements of CIFS affect the robustness enhancement, such as the top-$k$ feedback and architectures of the probe network.

\def \SubFigWidth {0.22} 
\def \SubImgWidth {1}
\begin{figure*}[h!]
	\centering
	\begin{subfigure}{\SubFigWidth\linewidth}
    	\includegraphics[width=\SubImgWidth \linewidth]{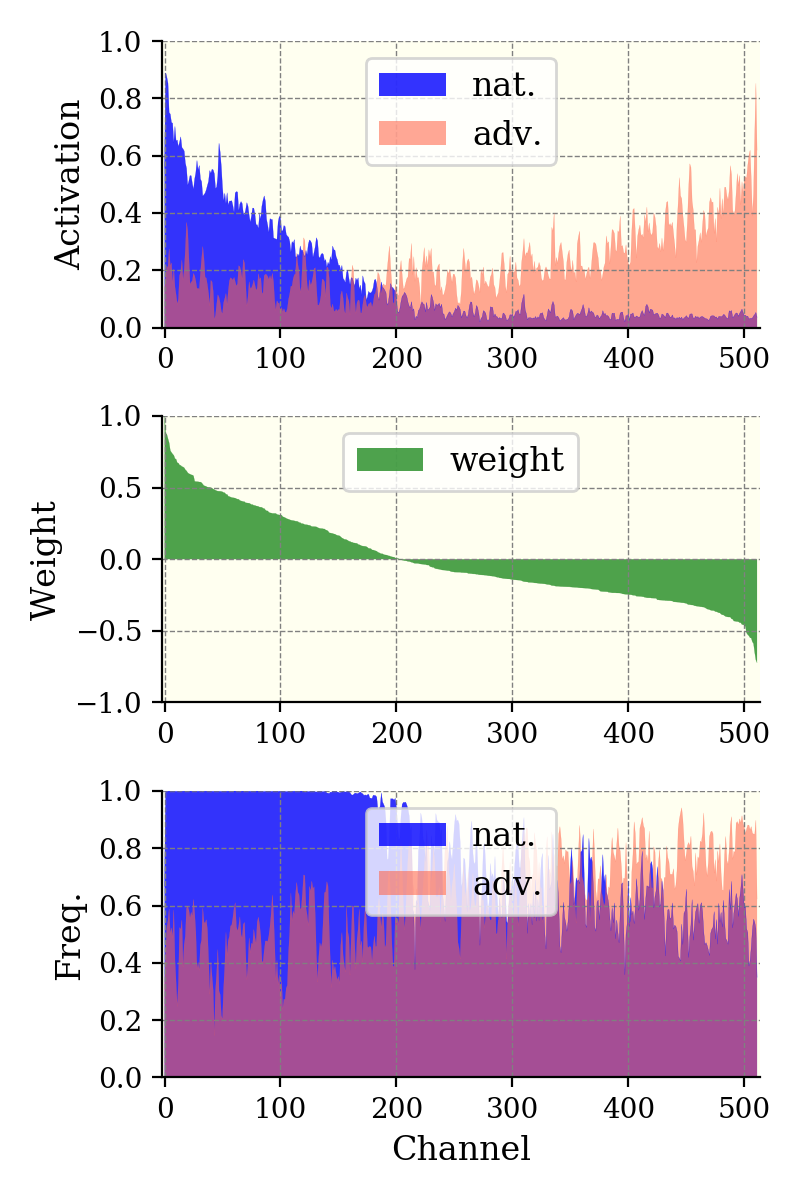}
    	\vspace{-2em}
    	\caption{{\scriptsize Normal / ``automobile''}}
    	\label{fig: w-vs-act_normal-auto}
	\end{subfigure}
	\begin{subfigure}{\SubFigWidth\linewidth}
    	\includegraphics[width=\SubImgWidth \linewidth]{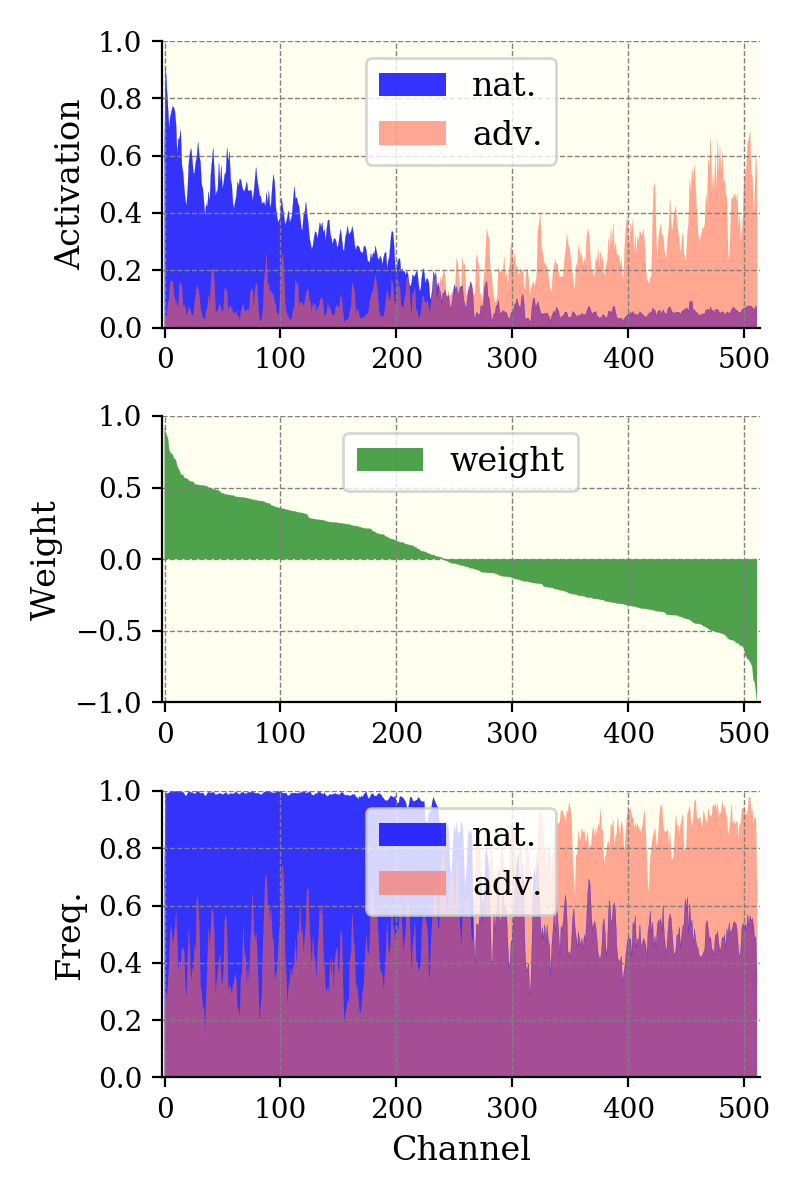} \vspace{-2em}
    	\caption{\scriptsize Normal / ``cat''}
    	\label{fig: w-vs-act_normal-cat}
	\end{subfigure}
	\begin{subfigure}{\SubFigWidth\linewidth}
    	\includegraphics[width=\SubImgWidth \linewidth]{images/hyp/PAT_Res18_cifar10___sorted_weight_vs_act_automobile}
    	\vspace{-2em}
    	\caption{\scriptsize Adv. / ``automobile''} 
    	\label{fig: w-vs-act_adv-auto}
	\end{subfigure}
	\begin{subfigure}{\SubFigWidth\linewidth}
    	\includegraphics[width=\SubImgWidth \linewidth]{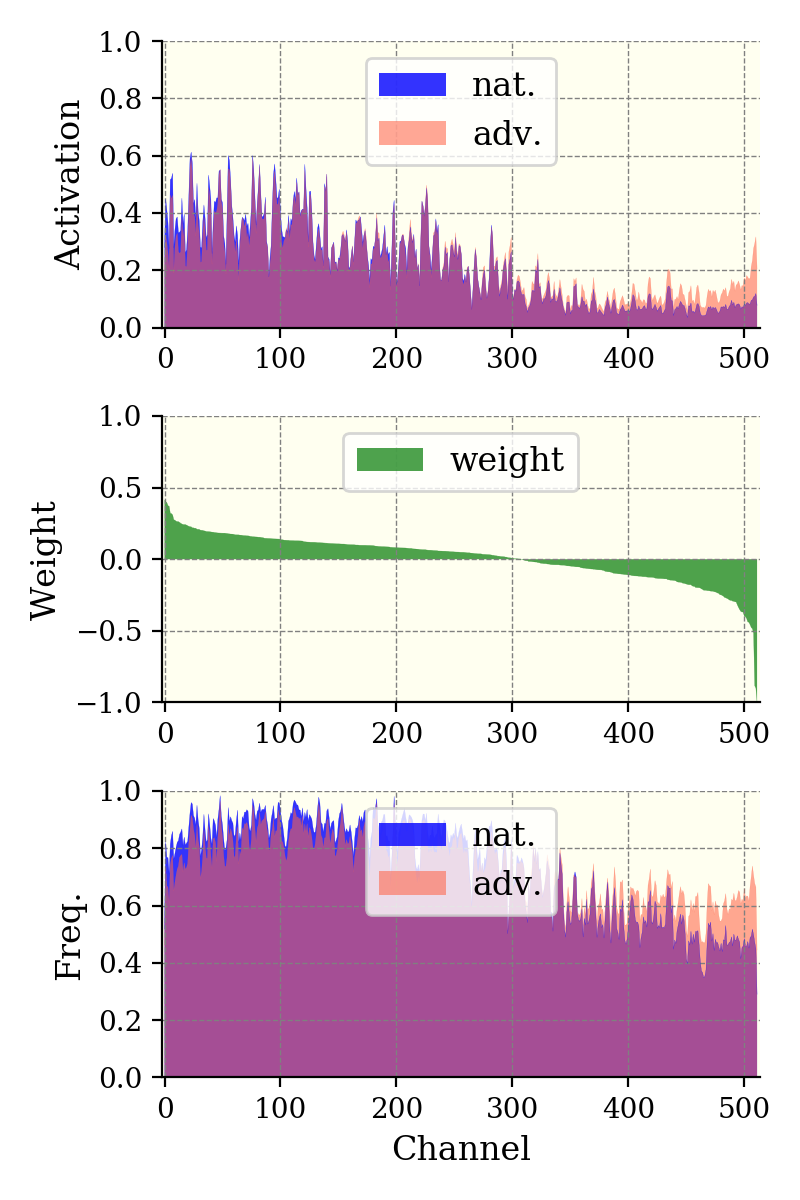}
    	\vspace{-2em}
    	\caption{\scriptsize Adv. / ``cat''}
    	\label{fig: w-vs-act_adv-cat}
	\end{subfigure}
    \vspace{-.5em}
    \caption{The magnitudes of channel-wise activations (top) at the penultimate layer, their activated frequencies (bottom), and the weights of the last linear layer (middle) \textit{vs.} channel indices. \#/$*$ means the CNN is trained in the \# way and we plot the average activations of data from the class $*$.
    \tit{The weights corresponding to class $*$ are sorted in the descending order and can indicate channels' relevances to class} $*$.
    The activation magnitudes and the activated frequencies of natural and adversarial (PGD-20) data are plotted according to the indices of the sorted weights.
     The robust accuracies on the whole CIFAR10 are 0\% for the normally trained ResNet-18 model and 46.6\% for the adversarially trained one (69.0\% for the ``automobile'' class and 16.7\% for the ``cat'' class). }
    \label{fig: w-vs-act}
    \vspace{-1em}
\end{figure*}

\section{Related Works}
This section briefly reviews relevant adversarial defense methods from two perspectives: adversarial training (AT)-based defense and robust network architecture design.

\paragraph{AT-based Defense} Adversarial training (AT) defends against adversarial attacks by utilizing adversarially generated data in model training \cite{goodfellow_explaining_2015}, formulated as a minimax optimization problem. Recently variants of AT \cite{cai_curriculum_2018,Wang_Xingjun_MA_FOSC_DAT,wang_improving_2020,wu2020adversarial,zhang_geometry-aware_2020} have been proposed. For example, the Misclassification-Aware-AdveRsarial-Training \cite{wang_improving_2020} modifies the process of generating adversarial data by simultaneously applying the misclassified natural data, together with the adversarial data for model training. Recent works have shown AT robustifies CNNs but degrades the natural accuracy \cite{tsipras_robustness_2018, zhang_theoretically_2019, lamb_interpolated_2019}. To achieve a better trade-off, \citet{zhang_theoretically_2019} decomposed the adversarial prediction error into the natural error and boundary error and proposed TRADES to control both terms at the same time. Besides, inspired by curriculum learning \cite{cai_curriculum_2018, bengio_curriculum_2009}, \citet{zhang_attacks_2020} proposed FAT to train models with increasingly adversarial data, which enhances generalization without sacrificing robustness. 

In addition, some works introduced various types of regularization for training models, such as layer-wise feature matching \cite{sankaranarayanan_regularizing_2018, liao_defense_2018, kannan_adversarial_2018}, low-rank representations \cite{sanyal_robustness_2020, mustafa_adversarial_2019}, attention map alignment \cite{xu_interpreting_2019}, and Lipschitz regularity\cite{Virmaux_lipschitz_2018, cisse_parseval_2017}. These types of regularization can work in conjunction with AT and benefit the models' robustness. 

\paragraph{Robust Network Design} Other than robust training strategies, some works explored robust network architectures \cite{yan_robustness_2020, hsieh_robustness_2019}. For instance, the work by \citet{yan_robustness_2020} showed neural ODE-based models are inherently more robust than conventional CNN models;  \citet{guo_sparse_2018} demonstrated that appropriately designed higher model sparsity implies better robustness of nonlinear networks. Another line of works defended against adversarial attacks via gradient obfuscation, such as random or non-differentiable image/feature transformations \cite{xie_mitigating_2018, du_rain_2020, dhillon_stochastic_2018, xiao_enhancing_2019}. However, they have been shown to be insecure to adaptive attacks \cite{athalye_obfuscated_2018, tramer_adaptive_2020}. Recently, many researchers have attempted to develop novel mechanisms for robustness enhancement. \citet{xie_feature_2019} performed feature denoising to remove the adversarial effects on feature maps. \citet{zoran_towards_2020} utilized the spatial attention mechanism to identify highlight important regions of feature maps. Most of these works manipulated CNNs' intermediate representations in the \tit{spatial domain}, whereas our work studies the adversarial robustness from the channel-wise activation perspective.

Channel-wise Activation Suppressing (CAS) \cite{bai_improving_2021}, the most relevant work to ours, also studied the channel-wise activations of adversarial data. It showed channels are activated more uniformly by adversarial data compared to the natural ones, and AT improves the robustness by attempting to align the distributions of channels' activations of natural and adversarial data. However, there are still some channels that are over-activated by adversarial data. To suppress these channels, the authors proposed CAS to adjust channels' activations based on their importance. Although CAS empirically suppresses certain channels, the authors did not show that the suppressed channels correspond to the target ones; this means the primary objective of CAS may not have been met. Thus, there is no guarantee CAS can enhance the robustness of CNNs (see Section \ref{sec:exp-eva} for further evidence on this). \tit{In contrast}, our work first builds a connection between robustness and channels' activations via their relevances to predictions. Then, the proposed CIFS can explicitly control channels' activations based on their relevances. Finally, experiments demonstrate the effectiveness of CIFS on robustness enhancement. 

\section{Channel-wise Importance-based Feature Selection}

In this section, we first study the adversarial robustness by comparing channels' activations of non-robust (normally trained) and robustified (adversarially trained) CNNs. Based on our observations of AT's effects, we propose a hypothesis on robustness enhancement via the adjustment of channels' activations (Section \ref{sec:cifs-adv-effects}). To examine this hypothesis, we then develop a novel mechanism, CIFS (Section \ref{sec:cifs-mechanism}), to manipulate channels' activation levels according to their relevances to predictions . Finally, we verify the proposed hypothesis through extensive experiments (Section \ref{sec:cifs-verify}). 

\subsection{Non-robust CNNs vs. Robustified CNNs: a Channel-wise Activation Perspective} \label{sec:cifs-adv-effects}

 
We compare a non-robust ResNet-18 \cite{he_deep_2016} model with an AT-robustified one on the CIFAR10 dataset \cite{krizhevsky_learning_nodate}. 
In ResNet-18, the representations of penultimate layer are spatially averaged   for each channel, then fed into the last linear layer for making predictions. Thus, the weights of the last linear layer indicate channels' relevances to predictions (according to the definition of channels' relevances in Introduction).
We visualize the channel-wise activation magnitudes, the activated frequencies (counted via a threshold of 1\% of the largest magnitude among all channels) in the penultimate layer for both natural and adversarial data, as well as the weights of the last linear layer in Figure~\ref{fig: w-vs-act}. The details of implementation are provided in Appendix~\ref{apdx:visualization-normal-at}.

From Figures~\ref{fig: w-vs-act_normal-auto} and \ref{fig: w-vs-act_normal-cat}, we observe, for a non-robust ResNet-18, the activation distribution of the adversarial data is obviously mismatched with that of the natural data: natural data activate channels that are PR to predictions with high values and high frequency, while adversarial data tend to amplify the NR ones.
From Figures~\ref{fig: w-vs-act_adv-auto} and \ref{fig: w-vs-act_adv-cat}, we observe that AT robustifies the model by aligning the activation distribution of adversarial data with that of natural data. Specifically, when dealing with adversarial data, AT boosts the activation magnitudes of PR channels while suppressing the activations of NR ones. However, we observe that, for many NR channels (e.g., around 150 channels from $350^{\text{th}}$ to $512^{\text{th}}$), the activations of adversarial data are much higher than those of natural data. These over-activations decrease the prediction scores corresponding to their true categories. Given this observation, we wonder \textcolor{OrangeRed}{(Q1)}: \textit{if we suppress these NR channels to regularize the freedom of adversarial perturbations, will it further improve the model's robustness upon AT}?

Besides, an adversarially trained model does not enjoy similar robustness for all classes, i.e., the robust accuracy of a certain class may be much higher than another (e.g., ``automobile'' with 69.0\% vs. ``cat'' with 16.7\% against PGD-20). Comparing the activations of these two classes (Figures~\ref{fig: w-vs-act_adv-auto} and \ref{fig: w-vs-act_adv-cat}), we observe that, for the class with strong robustness (e.g., ``automobile''), channels' activations align better with their relevances to labels, i.e., the channel with a greater extent of activation usually corresponds to a larger weight in the linear layer. In contrast, this alignment does not hold for the class with relatively poor robustness (e.g., for class ``cat'', the most activated channels, lying between the $26^{\text{th}}$ and $125^{\text{th}}$, are sub-PR to predictions).
Given this phenomenon, we may ask another question \textcolor{OrangeRed}{(Q2)}: \textit{{If we scale channels' activations based on their relevances to predictions, will it improve the model's robustness?}}

Considering the two questions above, we propose the unified hypothesis $\mathcal H$, as stated in the Introduction. 


\subsection{Importance-based Channel Adjustment} \label{sec:cifs-mechanism}

To examine the hypothesis $\mathcal H$, one needs a systematic approach to manipulate the channels, viz. selecting channels via suppressing NR ones but promoting PR ones. To this end, we introduce a mechanism, dubbed as Channel-wise Importance-based Feature Selection (CIFS). CIFS modifies layers of CNNs by adjusting channels' activations with importance scores that are generated from the channels' relevances to predictions. 

\begin{figure}[t!]
    \vspace{-.5em}
    \centering
    \includegraphics[width=1 \linewidth]{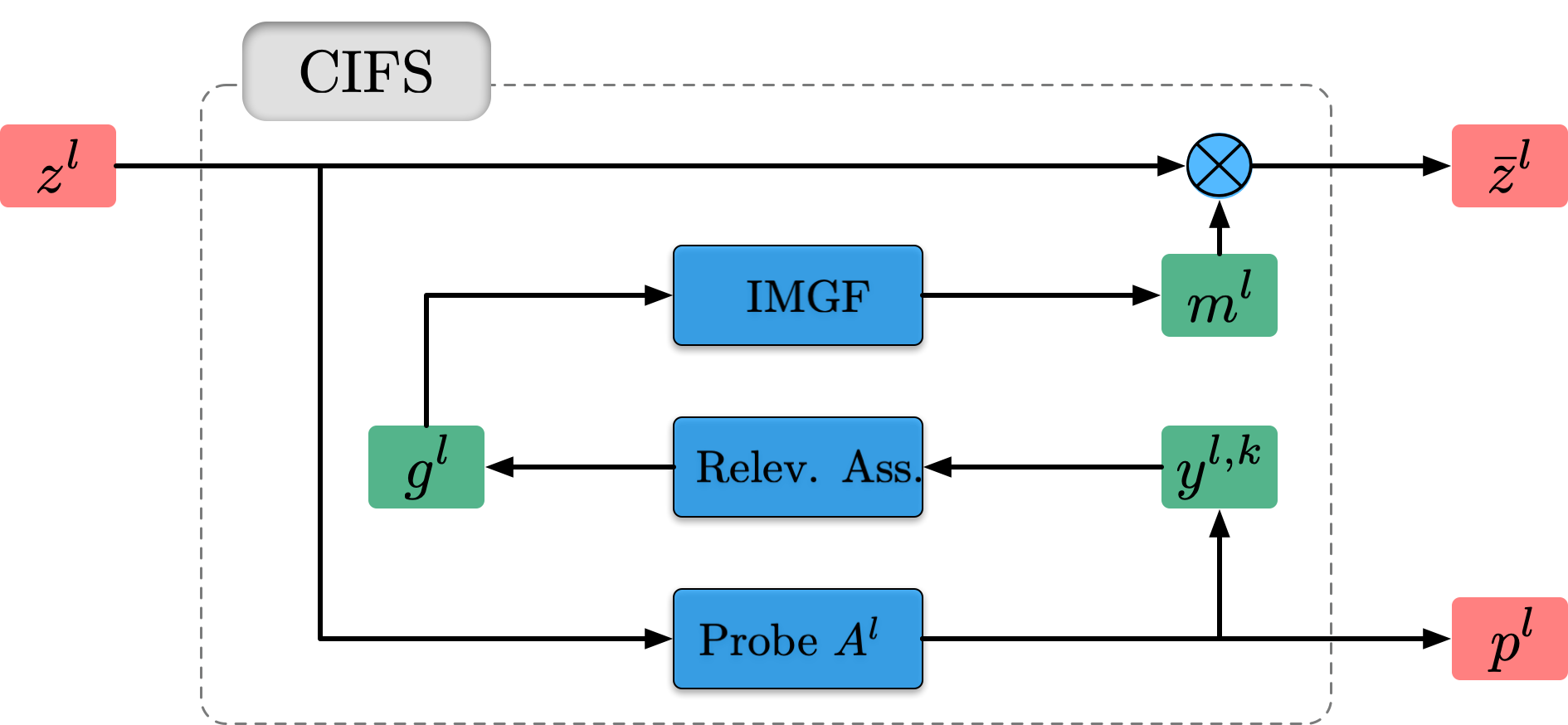}
    \vspace{-1.5em}
    \caption{CIFS: \tit{1)}~Probe Network $A^l$ first makes a raw prediction $p^l$ for $z^l$. \tit{2)}~Channels' relevances $g^l$ are assessed (Relev.~Ass.) based on the gradients of the top-$k$ prediction results $y^{l,k}$. \tit{3)}~The IMGF generates an importance mask $m^l$ from $g^l$ for channel adjustment.
    }
    \label{fig:CIFS}
    \vspace{-1.5em}
\end{figure}

For clearly state CIFS, we first introduce some notations: For a $K$-category classification problem, let $(X,Y) \sim P_{XY} $ denote the pair of the random input and its label, where $X\in \mathcal{X} \subset \sR^{n_X}$ and $Y\in \mathcal{Y} = \{0,1,\dots,K-1\}$. We design an $L$-layer CNN-based classification model to make accurate predictions for data sampled from $P_{XY}$. The $l^{\text{th}}$ layer is denoted by $f^{l}(\cdot)$ and parametrized by $\theta^{l} \in \Theta^l$; the mapping from the input to the $l^{\text{th}}$ layer's output is denoted by $f^{[l]}=f^{l}\circ f^{l-1}\circ \cdots \circ f^{1}$ and the combination of all the first $l$ layers' parameters is denoted by $\theta^{[l]}$, i.e., $\theta^{[l]}=(\theta^1, \dots, \theta^l)$.
Let us examine the $l^{\text{th}}$ layer where an input $x$ is transformed into a high-dimensional representation $z^l=f^{[l]}(x) \in \sR^{n^l_{\text{C}} \times n^l_{\text{F}}}$; $z^l$ has $n^l_{\text{C}}$ \underline{c}hannels and each channel is a \underline{f}eature vector of length $n^l_{\text{F}}$. With these notations, we elaborate the details of CIFS in \tit{\tbf{{three}}} steps (as shown in Figure \ref{fig:CIFS}).

\textcolor{Blue}{1) Surrogate Raw Prediction:}~To assess channels' relevances to predictions, 
a naive strategy is to compute the gradients of the final prediction with respect to $z^l$, i.e., $\nabla_{z^l} f^{[l+1:L]}(z^l)$, where $f^{[l+1:L]}=f^L \circ \cdots \circ f^{l+1}$. Since we need to adjust $z^l$ with importance scores generated from $\nabla_{z^l} f^{[l+1:L]}(z^l)$ and send the adjusted feature $\bar z^l$ to $f^{[l+1:L]}$ again for making the final prediction, it will result in computing the second-order derivatives during the training phase. Moreover, in practice, we may apply CIFS into multiple layers,
the forward pass will involve at least the second-order gradients (the latter CIFS-modified layer is recursively called). Thus, the back-propagation has to deal with at least the third-order gradients during training. This will aggravate the problem of training instability. 

Instead, inspired by the design of auxiliary classifiers in CAS \cite{bai_improving_2021}, CIFS builds a probe network $A^l$ as the surrogate of $f^{[l+1:L]}$ for a making raw prediction $p^l = A^l(z^l)$, so that we can use the gradients of $p^l$ to approximately assess the channels' relevances to the final prediction. The assessment does not involve other CIFS-modified layers. Thus, we can avoid the problem of back-propagation through high-order derivatives. The probe network $A^l$ is parameterized by $\theta^l_A$ and $p^l\in \sR^K$ represents the vector of prediction scores/logits. We can jointly optimize $A^l$ with the backbone network during the training phase under the supervision of true labels. 

\textcolor{Blue}{2) Relevance Assessment:}~
With the prediction $p^l$, we can compute the gradients of logits in $p^l$ \tit{w.r.t.} $z^l$ to assess the feature's relevances to each class.
We consider the top-$k$ prediction results ($k\geq 2$) for the assessment of channels' relevances. As data from two semantically similar classes (e.g., ``dog'' and ``cat'') usually share common features, the prediction for an input often assigns large scores to the classes similar to the true one and the top-$k$ results may include several of these similar classes \cite{jia_certified_2020}. In case the top-$1$ prediction is wrong, considering the top-$k$ results may help us reliably extract some common relevant features (see Section \ref{sec:exp-ablation} for more evidence). 

Let $y^{l,k}$ denote indices of the $k$ largest logits of prediction $p^l$. 
Let $\delta \in \sR^{n^l_{\text{C}}}$ be the channel-wise perturbation added to $z^l$, giving the perturbed representation $z^l_{\delta} = z^l + \delta \cdot \mathbf{1^{\top}}$. Here $\mathbf{1} \in \sR^{n^l_{\text{F}}}$ is the column vector with all elements as one, i.e., the features in the same channel are perturbed by a common value. We calculate the gradients of the sum of the top-$k$ logits \textit{w.r.t.} the channel-wise perturbation $\delta$:
\vspace{-0.5em}
\begin{equation}
 	\small
 	g^l 
 	= \left.\nabla_{\delta} \sum_{i \in y^{l,k}} p^l(\delta)_{[i]} \right\vert_{\delta=\mathbf{0}} 
 	= \left.\nabla_{z^l_{\delta}} \sum_{i \in y^{l,k}} A^l(z^l_{\delta})_{[i]} \right\vert_{z^l_{\delta} =z^l} \cdot \mathbf{1},
 	\label{eq:relevance-assessment}
\end{equation}
where $g^l=(g^l_{[0]},...,g^l_{[n^l_{\text{C}}-1]})$
represents the vector of channels' relevances to the top-$k$ logits. During the training phase, since the true label of $z^l$ is given,  we replace the top-$1$  prediction in $y^{l,k}$ with the true label $y$ and keep other prediction results untouched. 
	
\textcolor{Blue}{3) Importance Mask Generation:}~
As we want to suppress or promote channels based on their relevances, we need to design proper Importance Mask Generating Functions (IMGFs), which monotonically map relevances to \textit{non-negative} importance scores; of particular importance is to map negative relevances to values close to zero.

Here, we provide several feasible options: To answer the first question on whether suppressing NR channels enhances robustness, one can use the sigmoid function as the IMGF. With a large value of $\alpha$, the sigmoid function serves as a switch by mapping negative relevances to importance scores  close to zero and the positive close to one. 
To answer the second question concerning aligning channel activations with their relevances, we can use the softplus or softmax function as the IMGF. Both of them can map negative relevacnes to values close zero and map positive relevances monotonically to positive values.
These three functions are stated here for ease of reference:
\begin{itemize}
    \vspace{-.8em}
    \item sigmoid: $m^l_{[i]} = \frac{1}{1+\exp(-\alpha \cdot g^l_{[i]})}, \quad \alpha>0$. \vspace{-0.4em}
    \item softplus: $m^l_{[i]} = \frac{1}{\alpha} \cdot \log(1+\exp (\alpha \cdot g^l_{[i]})), \quad \alpha>0$. \vspace{-0.4em}
    \item softmax: $m^l_{[i]} = \frac{\exp(g^l_{[i]}/T)}{\sum_{j} \exp(g^l_{[j]}/T)}, \quad T>0$.
    \vspace{-.8em}
\end{itemize}
The usage of these functions will be discussed in detail in Section~\ref{sec:cifs-verify}. CIFS selects channels by multiplying the importance mask $m^l$ with $z^l$ as follows:
\begin{equation}
	\small \bar z^l = z^l \otimes \text{repmat}(m^l, 1, n^l_L),
\end{equation} 
where the ``$\text{repmat}$'' operation replicates the column vector $m^l$ along the second axis $n^l_L$ times.

\paragraph{Training of CIFS} In practice, we may apply the CIFS mechanism into several layers of a CNN. Let $I$ denote the set of indices of these layers, and $\theta_A^{I}$ denote the parameters of all the probes in the CIFS-modified layers. For each input $x$, the modified model $\bar f^{[L]}$ outputs $|I|$ raw predictions and one final prediction $p=\bar f^{[L]}(x)$. Given this, we use an adaptive loss function \cite{bai_improving_2021} for training the model: 
\begin{equation}
	\small
	\ell_{\beta}(x,y) = \frac{1}{1+\beta} \cdot \ell_{\mathrm{ce}}(p,y) + \frac{\beta}{(1+\beta)|I|} \cdot {\sum\limits_{l\in I} \ell_{\mathrm{ce}}(p^l, y)},
	\label{eq:adaptive-loss}
	\vspace{-1em}
\end{equation}
where $\ell_{\text{ce}}(\cdot, \cdot)$ denotes the cross-entropy loss and the coefficient $\beta>0$ balances the accuracy of raw predictions by CIFS and the final prediction. Since the subsequent decisions closely depend on the channels of features selected by the previous CIFS-modified layers, we should choose a proper value of $\beta$ to make sure that the raw predictions made by CIFS are reliable. In practice, we set $\beta$ to be $|I|$, and the effect of $\beta$ is discussed in the ablation study (see Appendix \ref{apdx:ablation}). To robustify the CNN model against malicious attacks, we can train $\bar f^{[L]}$ in an adversarial manner with a perturbation budget $\epsilon$. Namely, we solve the following optimization problem:  
\begin{equation}
	\small
	\min_{\theta^{[L]}, \theta^{I}_A} \E_{P_{XY}} \left[ \max_{X'\in \mathcal{B}(X, \epsilon, l_{\infty})} \ell_{\beta}(X',Y) \right],
	\label{eq:adv_cifs}
	\vspace{-.5em}
\end{equation}
where $\mathcal{B}(x, \epsilon, l_{\infty})=\{x' ~|~ \|x'-x\|_{l_{\infty}} \leq \epsilon \}$.

\def \SubFigWidth {0.22} 
\def \SubImgWidth {1}
\begin{figure*}[t!]
    \centering
    \begin{subfigure}{\SubFigWidth\linewidth}
        \centering
        \includegraphics[width=\SubImgWidth \linewidth]{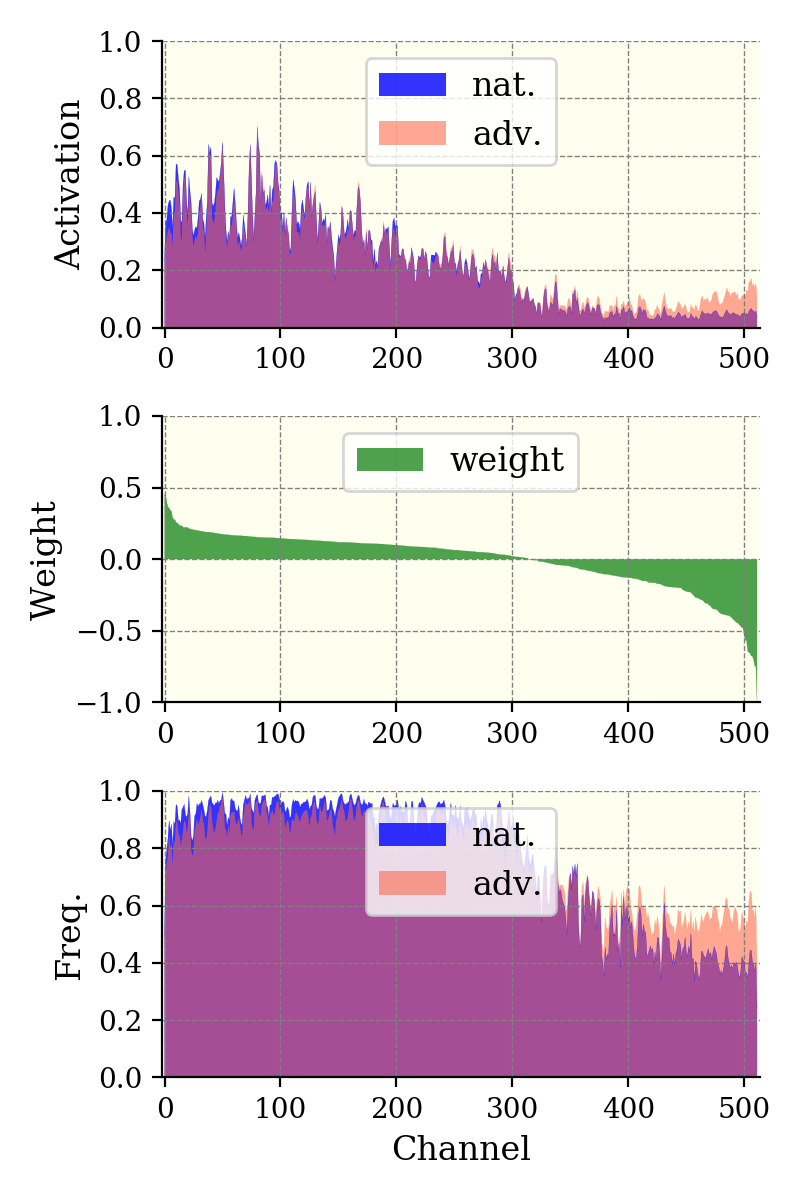}
        \vspace{-2em}
        \caption{\scriptsize non-CIFS}
        \label{fig:hyp-1-non-CIFS}
    \end{subfigure}
    \begin{subfigure}{\SubFigWidth\linewidth}
        \centering
        \includegraphics[width=\SubImgWidth \linewidth]{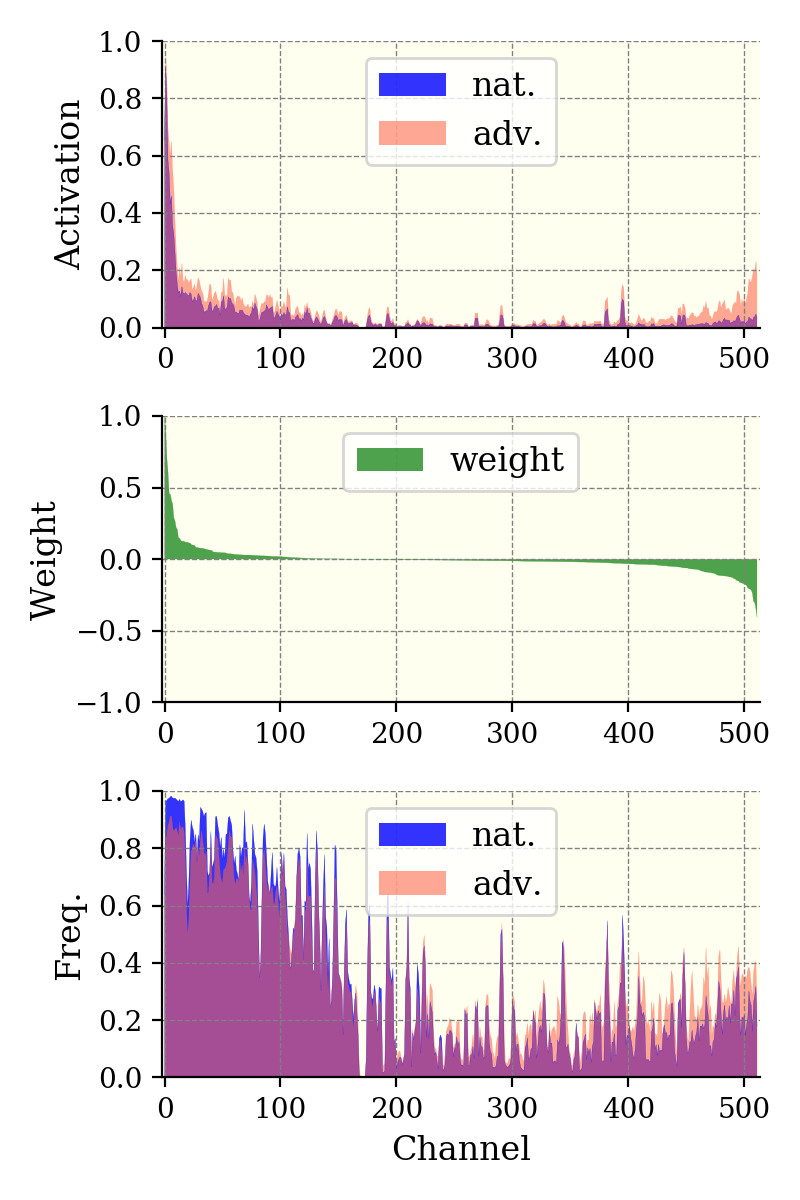}
        \vspace{-2em}
        \caption{\scriptsize CIFS-sigmoid}
        \label{fig:hyp-1-sigmoid}
    \end{subfigure}
    \begin{subfigure}{\SubFigWidth\linewidth}
        \centering
        \includegraphics[width=\SubImgWidth \linewidth]{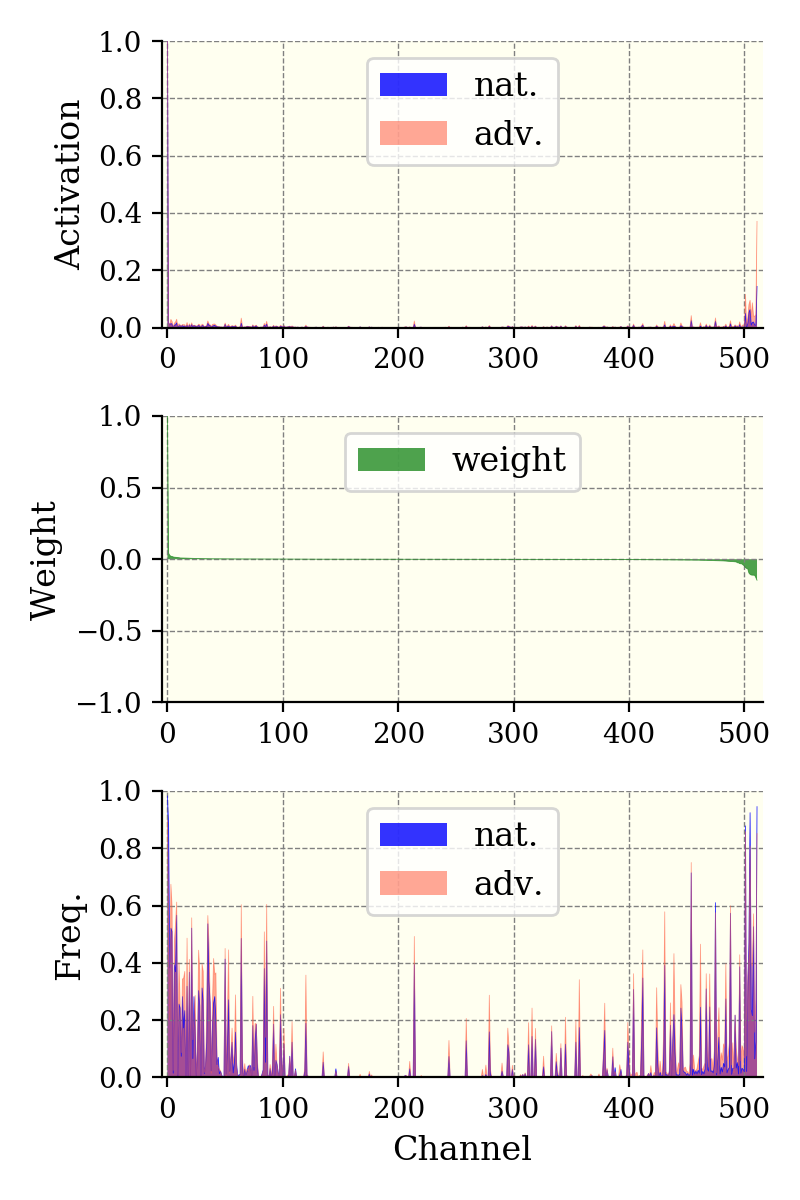}
        \vspace{-2em}
        \caption{\scriptsize CIFS-softplus}
        \label{fig:hyp-2-softplus}
    \end{subfigure}
    \begin{subfigure}{\SubFigWidth\linewidth}
        \centering
        \includegraphics[width=\SubImgWidth \linewidth]{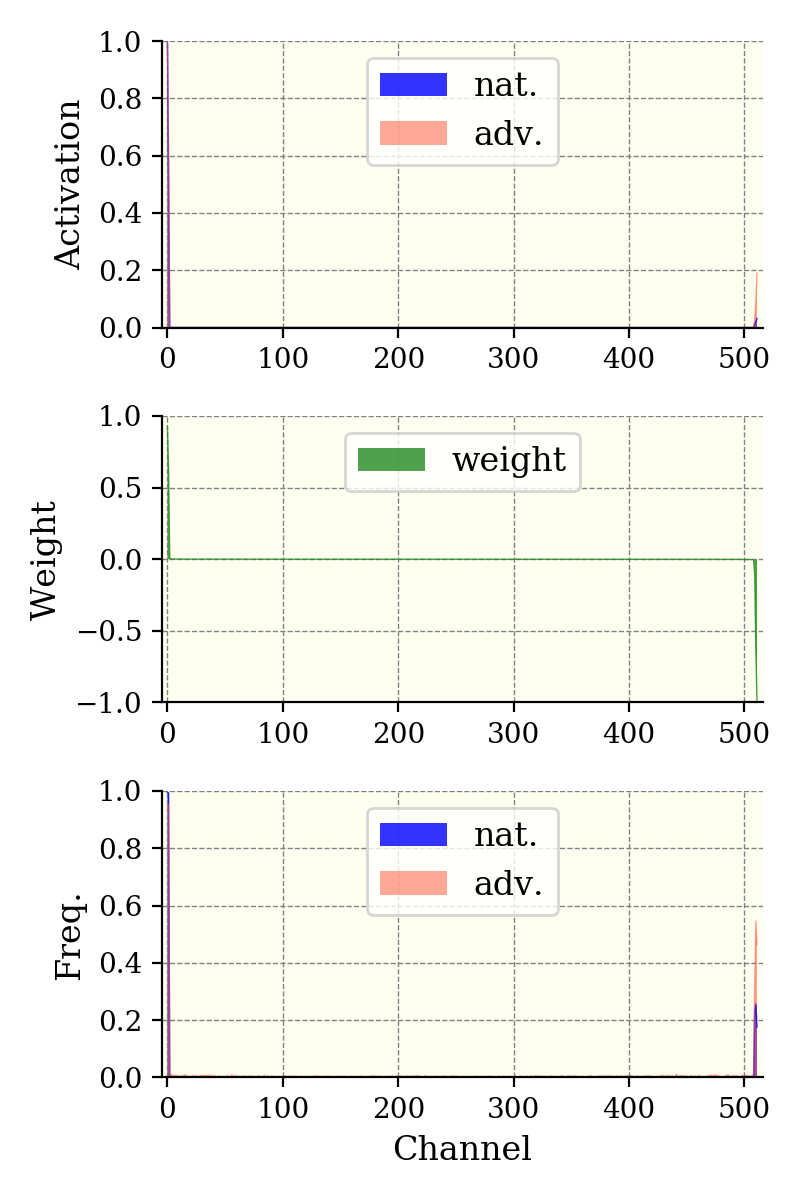}
        \vspace{-2em}
        \caption{\scriptsize CIFS-softmax}
        \label{fig:hyp-2-softmax}
    \end{subfigure}
    \vspace{-.5em}
    \caption{The magnitudes of channel-wise activations (top) at the penultimate layer, their activated frequency (bottom), and the weights of the last linear layer (middle) \textit{vs.} channel indices. The activations of natural data and their PGD-20 examples are averaged over all test samples in the ``airplane'' category. The robust accuracies against PGD-20 (on the whole dataset) are 46.64\% for non-CIFS, 49.87\% for the CIFS-sigmoid, 50.38\% for the CIFS-softplus, and 51.23\% for the CIFS-softmax respectively}
    \label{fig:hyp}
    \vspace{-1em}
\end{figure*}

\subsection{Verification of Hypothesis $\mathcal H$ on Robustness Enhancement} \label{sec:cifs-verify}

We verify the hypothesis $\mathcal H$ by answering the two questions, \textcolor{OrangeRed}{Q1} and \textcolor{OrangeRed}{Q2}, in Section~\ref{sec:cifs-adv-effects} respectively.


To answer \textcolor{OrangeRed}{Q1}, we applied the sigmoid function to generate the mask $m^l$ from $g^l$. Setting $\alpha$ to be large enough (here, $\alpha=10$), we can generate importance scores close to zeros for NR channels and scores close to one for the PR ones, so that we approximately annihilate the NR channels but leave the PR ones as unchanged. We adversarially trained a ResNet-18 model and its CIFS-modified version. As shown in Figure~\ref{fig:hyp-1-non-CIFS}, the NR channels (Channel 300-512) of the vanilla ResNet-18 model are clearly activated (the average activation magnitudes of these channels are larger than 0.1; the activation frequencies are over 0.4). In contrast, the ResNet-18 with CIFS-sigmoid effectively suppresses the activation of NR channels (Channel $100^{\text{th}}$--$512^{\text{th}}$ in Figure~\ref{fig:hyp-1-sigmoid}). Most of their mean activation magnitudes are smaller than 0.05, and their activation frequencies have clearly decreased. The experimental results show that, under AT, the vanilla ResNet-18 model results in a \underline{46.64\%} defense rate against the PGD-20 attack while its CIFS-sigmoid modified version achieves \underline{49.87\%}. More results can be found in Appendix \ref{apdx:visualization-cifs-modification}. Thus, we conclude that suppressing NR channels enhances the robustness of CNNs. 

To answer \textcolor{OrangeRed}{Q2}, we applied the softplus and softmax functions as IMGFs to generate the mask $m^l$ from $g^l$ respectively. Here the coefficient $\alpha$ in the softplus is set to be $5$ and the temperature $T$ in the softmax is set to be $1$. 
From Figures~\ref{fig:hyp-2-softplus} and \ref{fig:hyp-2-softmax}, we observe that, by generating importance scores positively correlated with the relevances, the model tends to completely focus on few relevant (positive and negative) channels. The channel of the greatest weight (most PR to predictions) is activated with the highest magnitude.
Most channels become irrelevant (small absolute values of weights) to the predictions and are activated at a low level. 
Using softplus as the IMGF (Figure~\ref{fig:hyp-2-softplus}), 
the irrelevant channels are sparsely activated, and the activation magnitudes are smaller than 5\% of the most important channel. 
In Figure~\ref{fig:hyp-2-softmax}, this phenomenon is enhanced by using softmax as IMGF: most channels become irrelevant to predictions and are usually deactivated. We evaluated the robustness of these two CIFS-modified CNNs against PGD-20 attack. Both of them outperformed the CIFS-sigmoid and the vanilla ResNet-18 classifiers (robust accuracies of CIFS-softplus and CIFS-softmax are \underline{50.38\%} and \underline{51.23\%} respectively, \tit{vs.},  \underline{49.87\%} for CIFS-sigmoid and \underline{46.64\%} for the vanilla ResNet-18). We also found CIFS can ameliorate the class-wise imbalance of adversarial robustness (\tit{e.g.}, CIFS-softmax increases the PGD-20 accuracy from 16.7\% to 22.3\% for class ``cat''). More details are provided in Appendix \ref{apdx:visualization-cifs-modification}. Thus, we conclude that aligning channels activations with their relevances to predictions can further robustify CNNs upon suppressing NR ones.

Given these empirical results, we verified the hypothesis $\mathcal H$ and justified that CIFS is an effective mechanism to improve the adversarial robustness of CNNs. In the following section, we conduct extensive experiments to evaluate the robustness enhancement through CIFS and study CIFS in an ablation manner.

\section{Experiments} \label{sec:exp}

\subsection{Robustness Evaluation} \label{sec:exp-eva}
We utilize the CIFS to modify CNNs in different architectures to perform classification tasks on benchmark datasets, namely a ResNet-18 and a WideResNet-28-10 on the CIFAR10 \cite{krizhevsky_learning_nodate} dataset, a ResNet-18 on the SVHN \cite{netzer_reading_nodate} dataset and 
a ResNet-10 on the Fashion-MNIST \cite{xiao2017/online} dataset.  We train the models with the standard PGD adversarial training (AT) \cite{madry_towards_2019} and its variants, such as FAT \cite{zhang_attacks_2020}, to show that CIFS can work under various AT-strategies.  We compare CIFS-modified CNNs with the vanilla versions as well as the CAS-modifications, where CAS \cite{bai_improving_2021} also modifies CNNs by adjusting channels' activations. Here, we report the results on CIFAR10 and SVHN. Results on FMNIST are presented in Appendix \ref{apdx:fmnist}.

\paragraph{Adaptive Attacks} As mentioned in Section \ref{sec:cifs-mechanism}, each CIFS-modified layer of CNNs outputs a raw prediction. To generate adversarial examples that are as strong as possible, we follow the strategy used in CAS and attack CNNs via the adaptive loss function $\ell_{\beta}$ in Equation (\ref{eq:adaptive-loss}) that considers all the raw and final predictions. We let the value of $\beta$ be chosen by the attacker, i.e., the attacker can try various values of $\beta_{\text{atk}}$ and select one that results in the most harmful perturbations. Our setting is more challenging for defense than CAS where the \tit{same} value of $\beta$ is used for training and attack. Here, for each adversarial attack, we evaluate the robustness by choosing $\beta_{\text{atk}}$ from $\{0, 0.1, 1, 2, 10, 100, \infty\}$ and report the worst robust accuracy\footnote{Results of various values of $\beta$ are present in Appendix \ref{apdx:cifar10-at}. We observe that CAS can improve the robustness of CNNs in most cases but \tit{fail} when attackers completely focus on CAS modules.}. Setting $\beta_{\text{atk}}=\infty$ means the attacks completely focus on the CIFS-modified layers \footnote{For $\beta_{\text{atk}}=\infty$, we consider the cases of attacking both CIFS-modified layers simultaneously and attacking each  separately.} and only consider the second term in Equation (\ref{eq:adaptive-loss}).

\subsubsection{Robustness Enhancement of CIFS under AT}
We adversarially train ResNet-18 and WRN-28-10 models with PGD-10 ($\epsilon=\nicefrac{8}{255}$) adversarial data.
CIFS is applied to the last two residual blocks of each model. The probes for the last and penultimate blocks are a linear layer and a multi-layer perceptron (MLP) respectively. Channels' relevances are assessed based on top-$2$ results and we use the softmax function with $T=1$ as the IMGF.
Other training details are provided in Appendix \ref{apdx:cifar10} and \ref{apdx:svhn}. 

\paragraph{Defense Results} We evaluate the robustness of CNNs against four types of white-box attacks: FGSM \cite{szegedy_intriguing_2014}, PGD-20 \cite{madry_towards_2019}, C\&W \cite{carlini_towards_2017}, and PGD-100. The $l_{\infty}$-norm of the perturbations are bounded by the value of $\epsilon=\nicefrac{8}{255}$. Here, we report the robustness evaluated at the last epoch for each model. Detailed attack settings and more defense results~( AutoAttack\footnote{AutoAttack consists of both white and black-box attacks.}~\cite{croce_reliable_2020} and the best epochs' results), are present in Appendix \ref{apdx:cifar10} and \ref{apdx:svhn}.

The defense results on CIFAR10 are reported in Table \ref{tab:white-box-cifar}. We observe that, for both of the ResNet-18 and WRN-28-10 architectures, CIFS consistently outperforms the counterparts against various types of adversarial attacks. For example, the CIFS-modified WRN-28-10 can defend the PGD100 attack with a success rate of $48.74\%$, which exceeds the second best by more than 4 percentage points. In contrast, under the strong adaptive attack, we see that the baseline CAS cannot improve and even worsens the robustness of CNNs. The defense results on SVHN are reported in Table \ref{tab:white-box-svhn}. The results also verify the  effectiveness of CIFS on improving robustness. 

\begin{table}[h]
    \vspace{-1em}
    \centering
    \caption{Robustness comparison of defense methods on CIFAR10. We report the accuracies (\%) for adversarial and natural data. For each model, the results of the strongest attack are marked with an underline.
    }
    \vspace{.5em}
    \scalebox{.8}{
    \begin{tabular}{cccccc}
    \toprule
    \tbf{\tit{ResNet-18}} & Natural & FGSM & PGD-20 & C\&W & PGD-100 \\
    \hline
    Vanilla & 84.56 & 55.11 & 46.62 & 45.95 & \underline{44.72} \\
    CAS     & 86.73 & 55.99 & 45.29 & 44.18 & \underline{43.22} \\
    CIFS    & 83.86 & \tbf{58.86} & \tbf{51.23} & \tbf{50.16} & \underline{\tbf{48.70}} \\
    \bottomrule
	\toprule 
    \tbf{\tit{WRN-28-10}} & Natural & FGSM & PGD-20 & C\&W & PGD-100 \\
    \hline
    Vanilla & 87.29 & 58.50 & 49.17 & 48.68 & \underline{47.08} \\
    CAS     & 88.05 & 57.94 & 49.03 & 47.97 & \underline{47.25} \\
    CIFS    & 85.56 & \tbf{61.34} & \tbf{53.74} & \tbf{53.20} & \underline{\tbf{51.51}} \\
    \bottomrule
    \end{tabular}
    }
    \label{tab:white-box-cifar}
    \vspace{-1em}
\end{table}{}

\begin{table}[h]
    \centering
    \caption{Robustness comparison of defense methods on SVHN. The accuracies (\%) for natural and adversarial data are reported. 
    }
    \vspace{.5em}
    \scalebox{.8}{
    \begin{tabular}{cccccc}
    \toprule
    \tbf{\tit{ResNet-18}} & Natural & FGSM & PGD-20 & C\&W & PGD-100 \\
    \hline
    Vanilla & 93.72 & \tbf{65.87} & 50.35 & 47.89 & \underline{45.81} \\
    CAS     & 94.08 & 65.24 & 48.47 & 46.15 & \underline{43.75} \\
    CIFS    & 93.94 & 66.24 & \tbf{52.02} & \tbf{50.13} & \underline{\tbf{47.49}} \\
    \bottomrule
    \end{tabular}
    }
    \label{tab:white-box-svhn}
    \vspace{-.5em}
\end{table}{}

\begin{figure}[h!]
    \vspace{-.3em}
	\centering
	\includegraphics[width=0.8\linewidth]{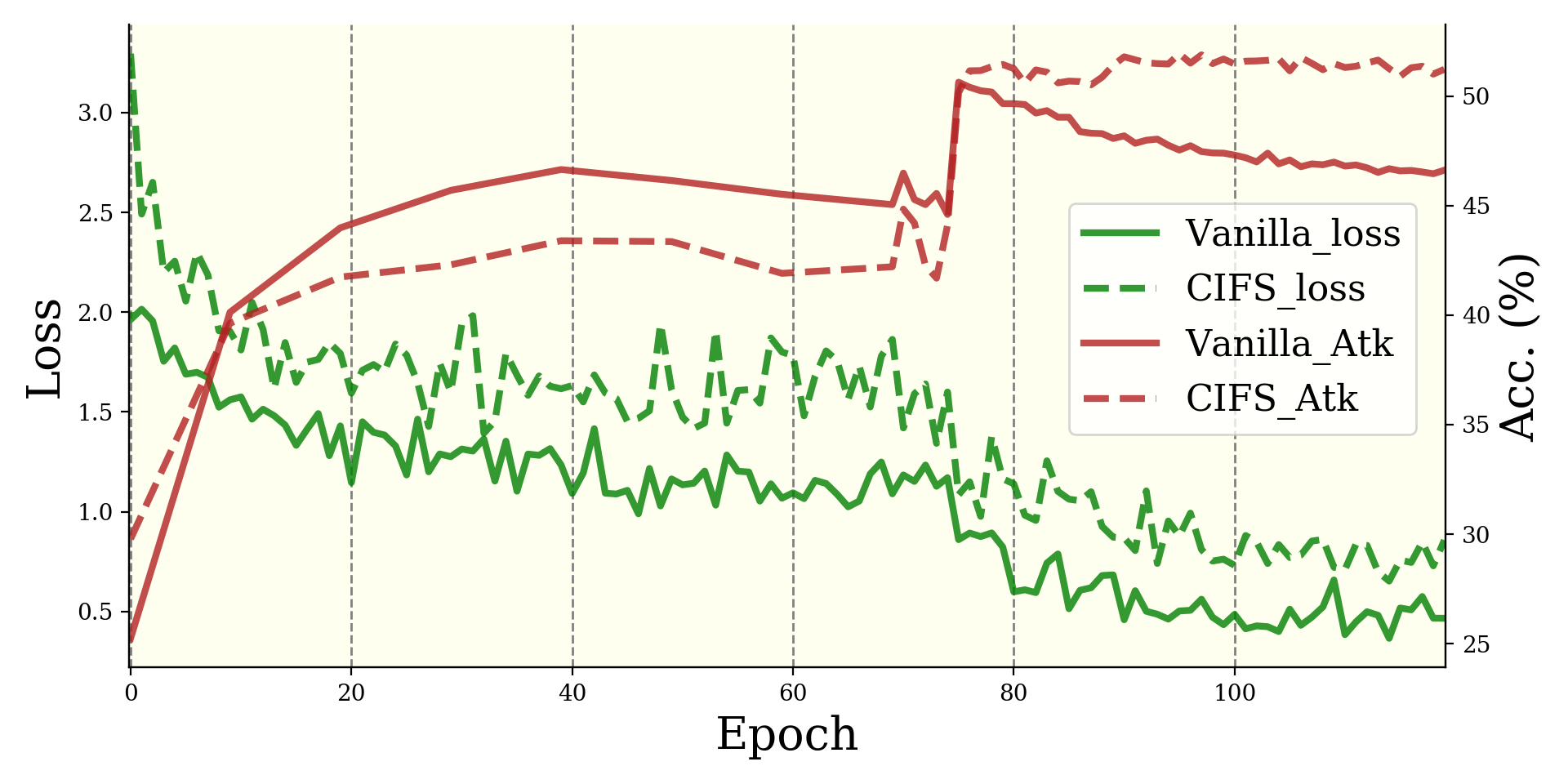}
	\vspace{-1em}
	\caption{Comparison on the training of a vanilla ResNet-18 and its CIFS-modified version. Training losses and accuracies against PGD-20 attack are plotted.}
    \label{fig:tr_procedure}
\end{figure}

\paragraph{Training Procedure} We train CNN classifiers in an adversarial manner for $120$ epochs and adjust the learning rate with a multiplier $0.1$ at epoch $75$ and epoch $90$. We summarize the training procedure by plotting the curves of training losses and the PGD-20 accuracies \tit{w.r.t.} epochs in Figure \ref{fig:tr_procedure}. We observe the best adversarial robustness of the vanilla ResNet-18 ($50.64\%$ defense rate) appears around the $75^{\text{th}}$ epoch. After epoch $75$, the model starts to overfit to training data, i.e., the training loss continues decreasing, but the robust accuracy drops as well. In contrast, the overfitting problem is ameliorated by the application of CIFS. We can see that the best robust accuracy appears around the $90^{\text{th}}$ epoch; After the best epoch, the training loss continues decreasing, but the robustness is maintained around the peak. This phenomenon may result from the fact that CIFS suppresses redundant channels. The model redundancy can be controlled by selecting few highly relevant channels and deactivating others. In this way, the overfitting in training is ameliorated.

On the computation overhead, we report the training time and the evaluation time of a ResNet-18 classifier on the CIFAR10 dataset for reference. For PGD-10 adversarial training, the vanilla CNN takes 166s for each epoch while the CIFS-modified model takes 172s. For the PGD-20 evaluation, the vanilla CNN takes 53s for the CIFAR10 test set; the CIFS-modified model takes 56s instead. In short, the proposed CIFS does not result in too much extra computation. 

\subsubsection{CIFS Working in Conjunction with Variants of AT}
CIFS improves the adversarial robustness by adjusting channels' activations, which is orthogonal to defense training strategies. Here, we train the CIFS-modified CNNs with variants of AT to examine whether CIFS can work in conjunction with other state-of-the-art training-based defense techniques. We consider the FAT \cite{zhang_attacks_2020} and TRADES \cite{zhang_theoretically_2019} strategies. 
We report the defense results of FAT in Table \ref{tab:fat} and provide the results of TRADES in Appendix \ref{apdx:cifar10-trades}. We observe that FAT training strategy improves the natural accuracy and robustness of CNNs (compared to the results in Table \ref{tab:white-box-cifar}). Under the FAT strategy, CIFS also improves the adversarial robustness of the vanilla CNNs in both ResNet-18 and WRN-28-10 architectures.

\begin{table}[h]
    \vspace{-1em}
    \centering
    \caption{Robustness comparison of defense methods trained with FAT on CIFAR10. Comparing the accuracies (\%) against the strongest attacks, we observe that CIFS clearly robustifies CNNs.}
    \vspace{.5em}
    \scalebox{.8}{
    \begin{tabular}{cccccc}
    \toprule
    \tbf{\tit{ResNet-18}} & Natural & FGSM & PGD-20 & C\&W & PGD-100 \\
    \hline
    Vanilla & 87.16 & 56.43 & 47.64 & 46.01 & \underline{45.35} \\
    CIFS     & 86.35 & \tbf{59.47} & \tbf{51.68} & \tbf{51.84} & \underline{\tbf{49.52}} \\
    \bottomrule
    \toprule
    \tbf{\tit{WRN-28-10}} & Natural & FGSM & PGD-20 & C\&W & PGD-100 \\
    \hline
    Vanilla & 88.37 & 58.81 & 49.62 & 48.49 & \underline{47.58}\\
    CIFS  & 86.74 & \tbf{60.67} & \tbf{51.99} & \tbf{52.34} & \underline{\tbf{49.87}}\\
    \bottomrule
    \end{tabular}
    }
    \label{tab:fat}
\end{table}{}

\subsection{Ablation Study}  \label{sec:exp-ablation}
Here, we conduct an ablation study to further understand the robustness properties of CIFS. Specifically, we investigate the effects of the feedback from the top-$k$ predictions. The ablation experiments are conducted on CIFAR10 based on the ResNet-18 model. Besides, in Appendix \ref{apdx:ablation}, we also study cases in which the CIFS is applied to different layers, the probe networks are in different architectures, various values of $\beta$ are used for training. 

\paragraph{Feedback from Top-$k$ Prediction Results}
As is well-known, the top-$k$ classification accuracy for $k>1$ is always not worse than the top-$1$. 
For example, in Table \ref{tab:topk}, we can see that the top-$2$ accuracy of an adversarially trained ResNet-18 against the PGD-20 attack exceeds the top-$1$ accuracy by $25$ percentage points. This implies that, although adversarial data can usually fool the classifier (i.e., low top-$1$ accuracy), the prediction confidence of the true class is still high and the corresponding score highly likely lies among the top-$2$ or $3$ logits.

\begin{table}[h]
	\vspace{-1em}
    \centering
    \caption{Top-$k$ accuracies (\%) against adversarial attacks on CIFAR10 of an adversarially trained ResNet-18.}
    \vspace{.5em}
    \scalebox{.8}{
    \begin{tabular}{cccccc}
    \toprule
    \tbf{\tit{ResNet-18}} & top-$1$ & top-$2$ & top-$3$\\
    \hline
    FGSM & 55.11 & 76.22 & 85.20 \\
    PGD-20 & 46.62 & 71.71 & 81.60 \\
    \bottomrule
    \end{tabular}
    }
    \label{tab:topk}
\end{table}{}

CIFS generates the importance mask from the raw prediction and uses it to suppress or promote channels at the current layer. The final prediction made by subsequent layers strongly depends on the channels selected by CIFS. To ensure the accuracy of final predictions, the logits used for generating importance scores should include the true label's logit for each input so that the truly important channels will be highlighted. According to Table \ref{tab:topk}, if we use the top-$1$ logit to assess the importance of channels for PGD-20 adversarial data, the probability of incorrect assessment is over $50$\%. Instead, if we use the feedback from top-$2$ or $3$ logits, the truly important channels can highly likely be promoted. The following table presents more experiments that justify this argument. 

\begin{table}[h]
	\vspace{-1em}
    \centering
    \caption{Robustness comparison (\%) of importance assessment based on top-$k$ results against the PGD-20 attack. The column header $*/\#$ represents the attack point and the output prediction. For example, CIFS/Final means the model outputs the final prediction and the attacker solely focuses on the CIFS's raw prediction.}
    \vspace{.5em}
    \scalebox{.8}{
    \begin{tabular}{cccccc}
    \toprule
    \tbf{\tit{ResNet-18}} & Natural/Final & CIFS/CIFS & CFIS/Final & Adap/Final\\
    \hline
    Vanilla & 84.56 & - & - & 46.62 \\
    top-$1$ & 87.63 & 47.24 & 47.24 & {47.24} \\
    top-$2$ & 83.86 & 48.72 & 54.96 & {\tbf{51.23}} \\
    top-$3$ & 83.49 & 47.59 & 55.39 & {49.91} \\
    \bottomrule
    \end{tabular}
    }
    \label{tab:purify}
\end{table}{}

From Table \ref{tab:purify}, we observe that, for the top-$1$ case, the defense rate of `CIFS/CIFS' is the same as that of `CIFS/Final' and that of `Adap/Final.' This implies that, once an adversarial example successfully fools the raw prediction of CIFS, the final prediction also will be incorrect. Thus, the attacker only needs to focus on the CIFS's raw predictions to break the model. In contrast, for the top-$2$ case, the defense rate of `CIFS/Final' exceeds the `CIFS/CIFS' by 6 percentage points. This means that nearly 6\% adversarial data mislead the CIFS's raw predictions. However, through the channel adjustment via CIFS, these adversarial data are ``purified'', and more relevant characteristic features are thus transmitted to subsequent layers of CNNs. As such, these adversarial data are finally classified correctly. In this case, the attacker has to exhaustively search for an adaptive loss function to generate attacks, and the CIFS-modified CNNs are safer and more reliable. More discussion on why the top-$k$ assessment performs better and how to choose $k$ is provided in Appendix \ref{apdx:ablation}.

\section{Conclusion}

We developed the CIFS mechanism to verify the hypothesis that suppressing NR channels and aligning PR ones with their relevances to predictions benefits adversarial robustness. Empirical results demonstrate the effectiveness of CIFS on enhancing CNNs' robustness.

There are two limitations of our current work: 1) We empirically verify the hypothesis $\mathcal H$, but it is still difficult to explicitly, not intuitively, explain why the adjustment of channels improves robustness. 2) Although CIFS ameliorates the overfitting during AT and improves the robustness, it sometimes leads to a bit drop in natural accuracies on certain datasets. In the future, we will attempt to address these two limitations.

\section*{Acknowledgements}
HY and VYFT are funded
by  a Singapore National Research Foundation (NRF) Fellowship
(R-263-000-D02-281).

JF is supported by the National Research Foundation Singapore under its AI Singapore Programme (Award Number: AISG-100E-2019-035)

JZ, GN, and MS are supported by JST AIP Acceleration Research Grant Number JPMJCR20U3, Japan. MS is also supported by the Institute for AI and Beyond, UTokyo.

\bibliography{_references_}
\bibliographystyle{icml2021.bst}


\clearpage
\appendix
\counterwithin{figure}{section}
\counterwithin{table}{section}

\section{Details on Visualizing Channel-wise Activations} \label{apdx:visualization}

\subsection{Non-robust CNNs vs. Robustified CNNs} \label{apdx:visualization-normal-at}
We train ResNet-18 models to perform the classification task on the CIFAR10 dataset. The models are trained normally and adversarially. We use adversarial data generated by PGD-10 attack ($\epsilon=\nicefrac{8}{255}$, step size $\nicefrac{\epsilon}{4}$, and random initialization) for adversarial training.

The ResNet-18 network consists of one convolutional layer, eight residual blocks, and one linear fully-connected (FC) layer connected successively. Each residual block contains two convolutional layers for the residual mapping. We visualize the features of the penultimate layer (the output of the eighth residual block) and the weights of the last linear layer in Figure \textcolor{Blue}{1}. Specifically, the weights of the last FC layer for a certain class are sorted and plotted in descending order. We process the penultimate layer's features with the global average pooling operation to obtain the channel-wise activations. For a certain class, we calculate each channel's mean activation magnitude over all the test samples in this category. We normalize the mean channel-wise activations by dividing them by their absolute maximum. The mean channel-wise activations are plotted according to the indices of the sorted weights. We also record the activated frequency of each channel. Here, the channel is regarded to be activated if its activation magnitude is larger than a threshold (1\% of the maximum of all channels' activations). 

\subsection{Channel-wise Activations of CIFS-modifed CNNs}
\label{apdx:visualization-cifs-modification}
We train CIFS-modified CNNs normally and adversarially by using the adaptive loss in Equation (\textcolor{Blue}{3}). We use the PGD-10 attack to generate adversarial data. We illustrate the channels' activations of CIFS-modified CNNs in Figure \textcolor{Blue}{3}. The implementation details are same as those in Appendix \ref{apdx:visualization-normal-at}.

In Figure \textcolor{Blue}{3}, we show the channels' activations of data in class ``airplane''. Here, we plot the channels activations of data in other classes. From Figure \ref{fig:apx-hyp}, we see that CIFS indeed suppresses negatively-relevant (NR) channels and promotes the positively-relevant (PR) ones.

\begin{figure}[h!]
	\centering
	\includegraphics[width=.9\linewidth]{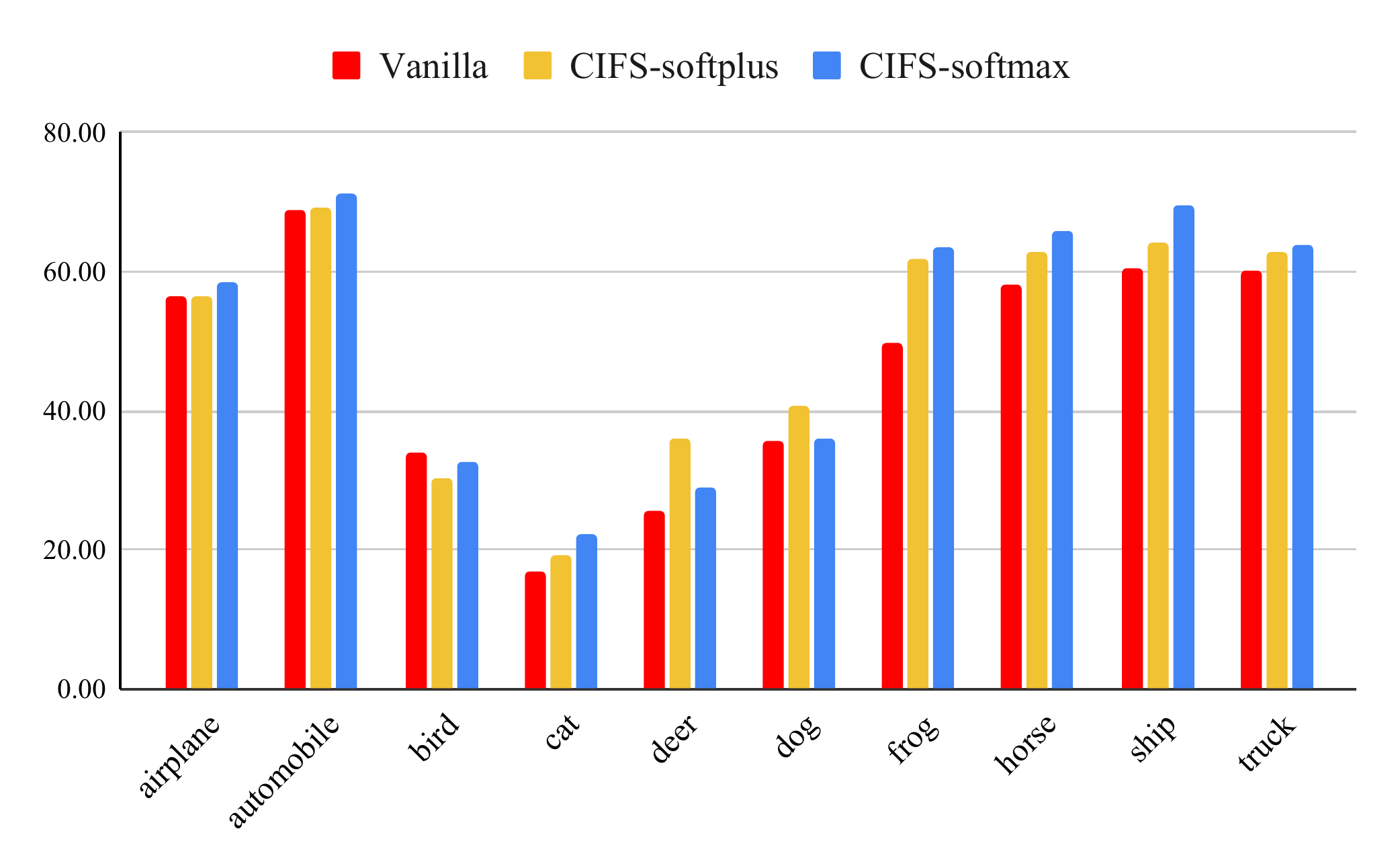}
	\vspace{-1em}
	\caption{Robust accuracies (\%) of PGD-20 adversarial data for various classes of the CIFAR10 dataset.}
	\label{fig:apdx-imbal}
\end{figure}

Besides, we also observe that CIFS ameliorates the class-wise imbalance of robustness under AT. In Figure \ref{fig:apdx-imbal}, we can see that, for the data in class ``cat'' and class ``deer'', the robust accuracies of the vanilla ResNet-18 model are 16.70\% and 25.50\%. Modifying the vanilla model with CIFS-softmax, we can improve the robust accuracies by 5.6 and 3.3 percentage points, respectively.

\def \SubFigWidth {0.245} 
\def \SubImgWidth {1}
\begin{figure*}[t!]
    \centering
    \begin{subfigure}{\SubFigWidth\linewidth}
        \centering
        \includegraphics[width=\SubImgWidth \linewidth]{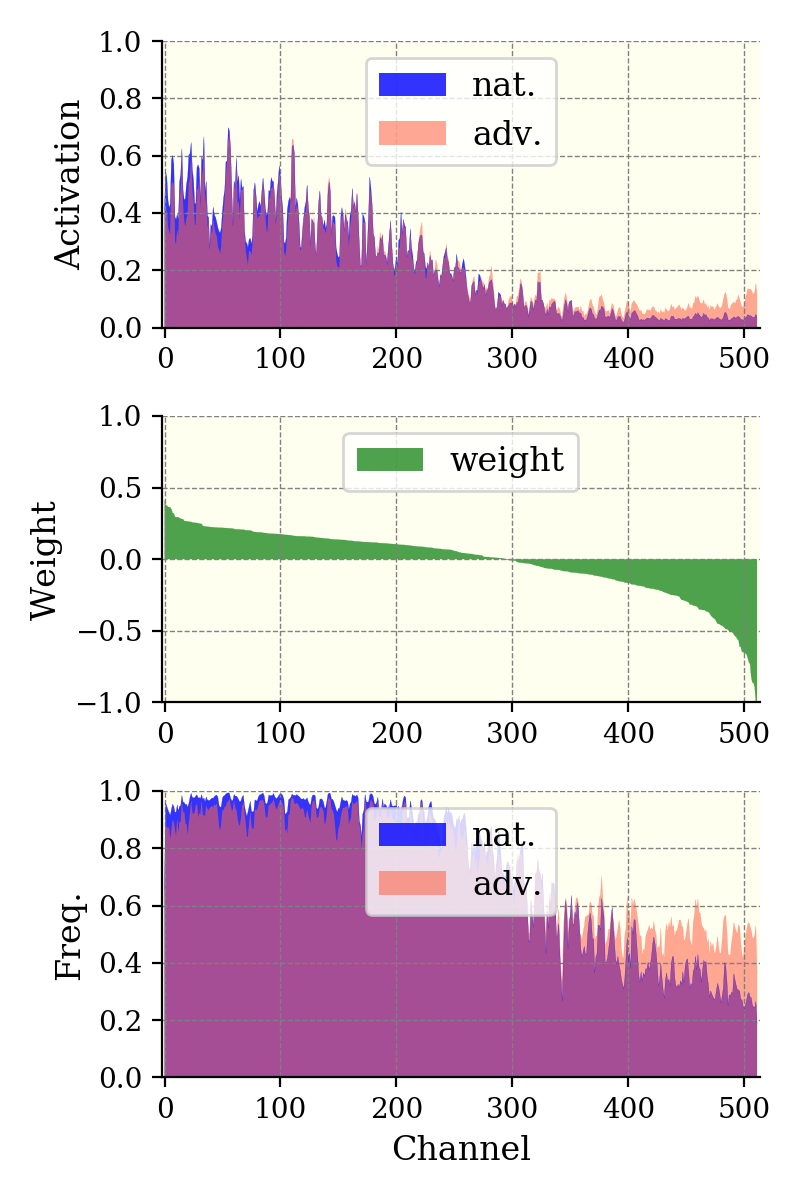}
        \vspace{-2em}
        \caption{\scriptsize ``ship'': non-CIFS}
    \end{subfigure}
    \begin{subfigure}{\SubFigWidth\linewidth}
        \centering
        \includegraphics[width=\SubImgWidth \linewidth]{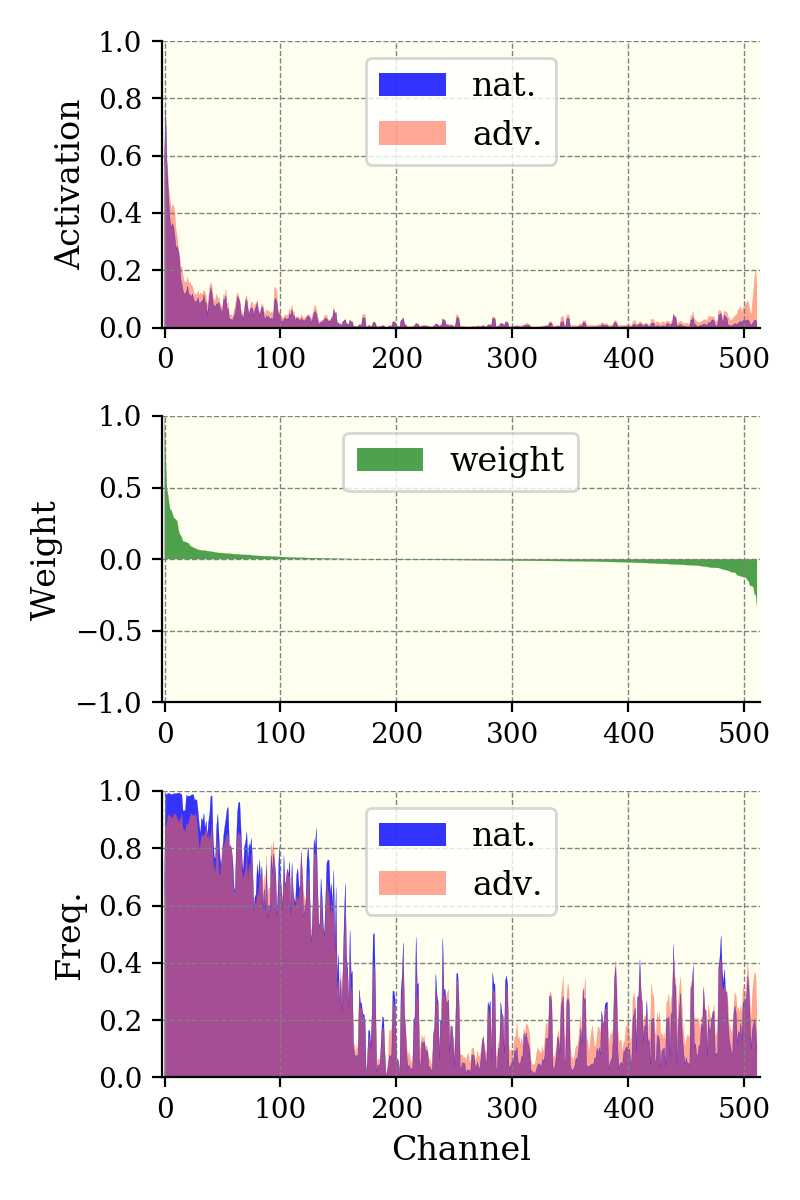}
        \vspace{-2em}
        \caption{\scriptsize ``ship'': CIFS-sigmoid}
    \end{subfigure}
    \begin{subfigure}{\SubFigWidth\linewidth}
        \centering
        \includegraphics[width=\SubImgWidth \linewidth]{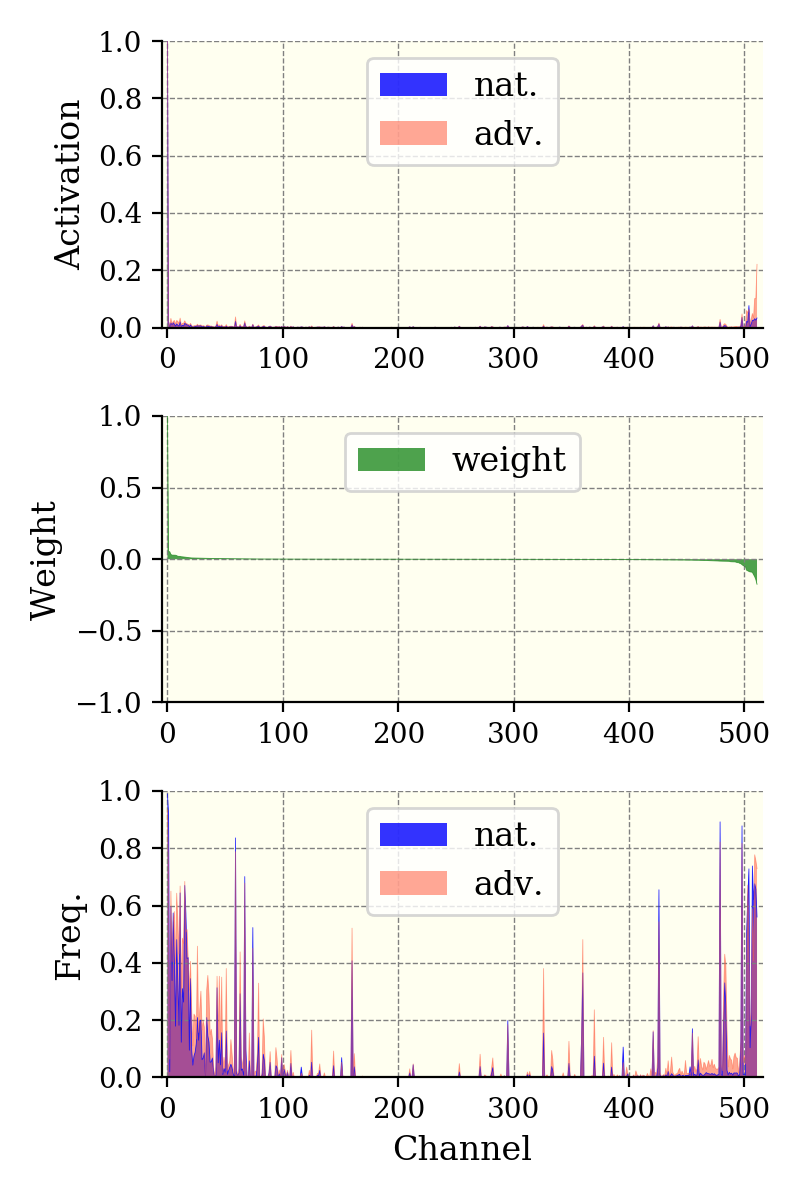}
        \vspace{-2em}
        \caption{\scriptsize ``ship'': CIFS-softplus}
    \end{subfigure}
    \begin{subfigure}{\SubFigWidth\linewidth}
        \centering
        \includegraphics[width=\SubImgWidth \linewidth]{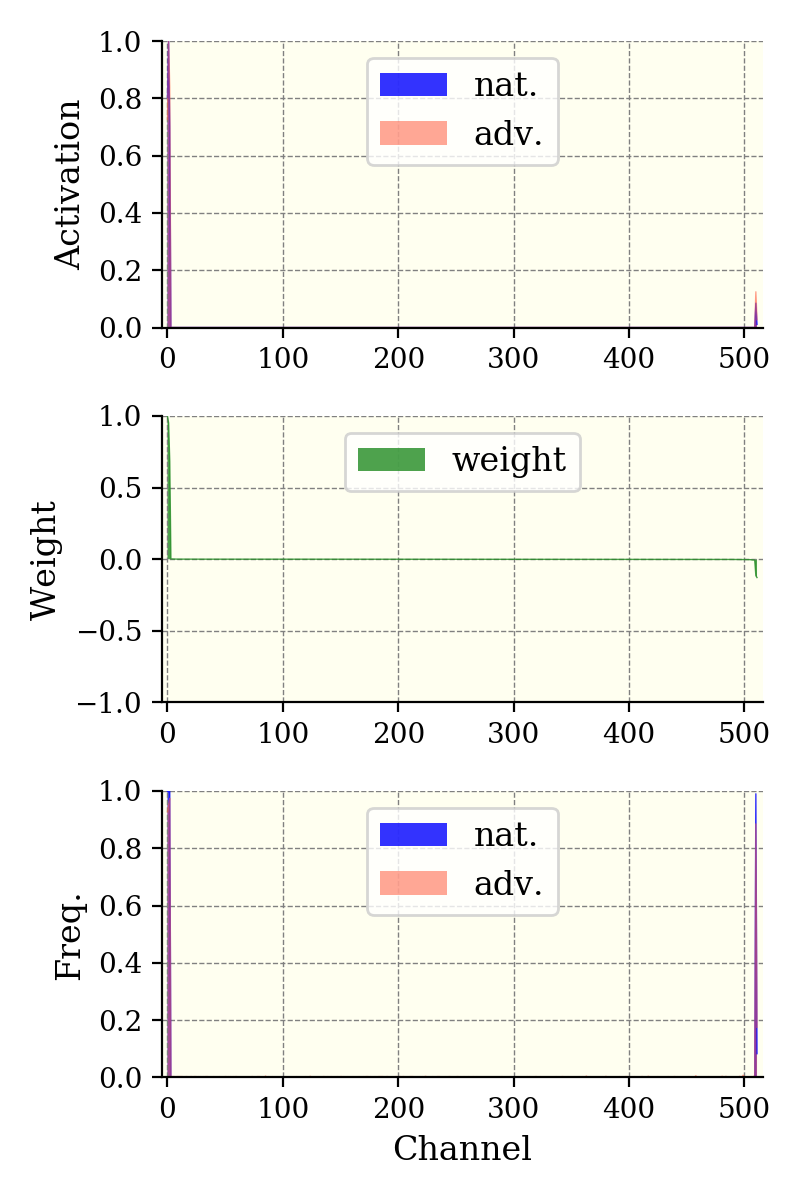}
        \vspace{-2em}
        \caption{\scriptsize ``ship'': CIFS-softmax}
    \end{subfigure}
    
    \begin{subfigure}{\SubFigWidth\linewidth}
        \centering
        \includegraphics[width=\SubImgWidth \linewidth]{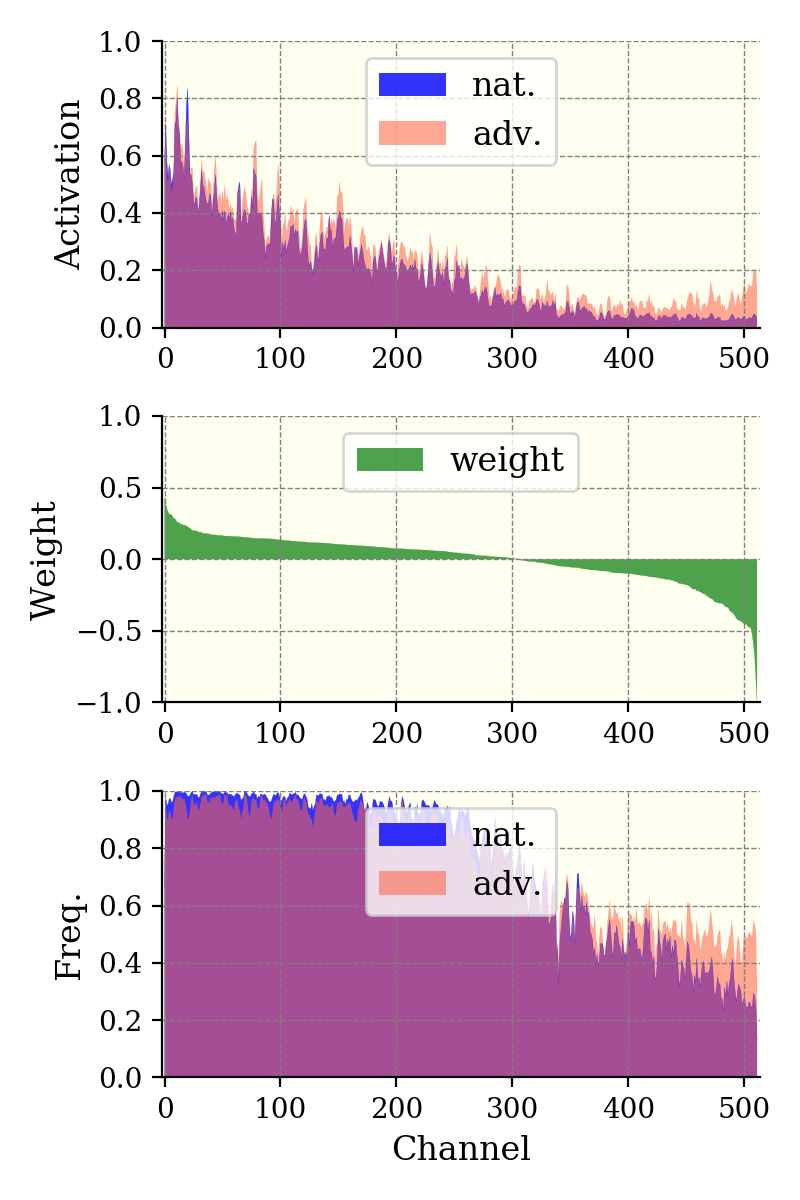}
        \vspace{-2em}
        \caption{\scriptsize ``autmobile'': non-CIFS}
    \end{subfigure}
    \begin{subfigure}{\SubFigWidth\linewidth}
        \centering
        \includegraphics[width=\SubImgWidth \linewidth]{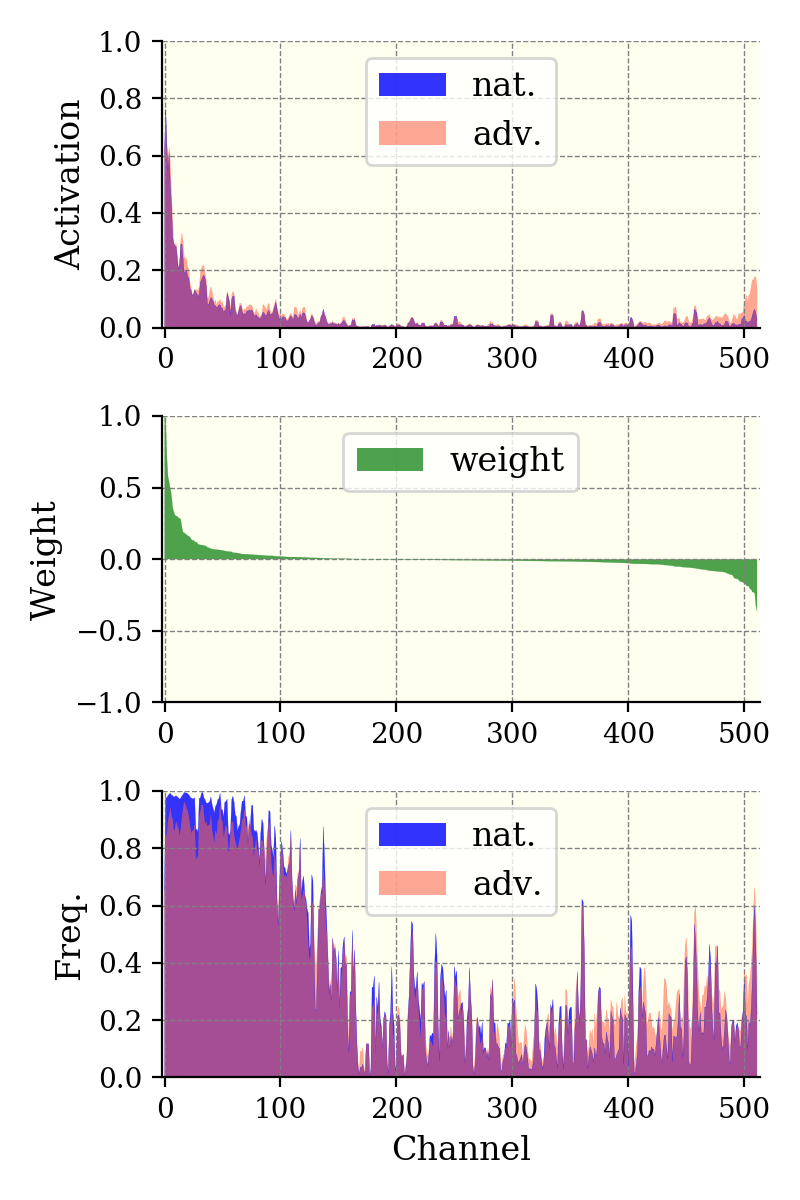}
        \vspace{-2em}
        \caption{\scriptsize ``autmobile'': CIFS-sigmoid}
    \end{subfigure}
    \begin{subfigure}{\SubFigWidth\linewidth}
        \centering
        \includegraphics[width=\SubImgWidth \linewidth]{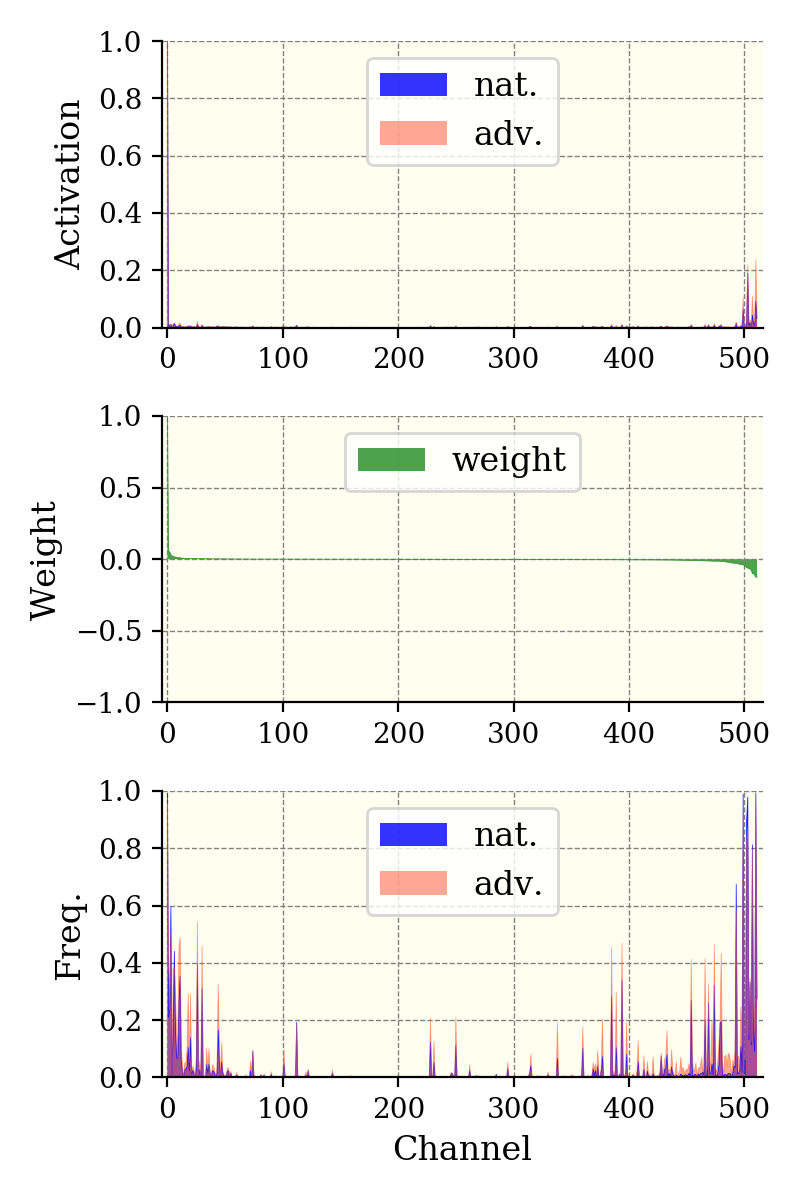}
        \vspace{-2em}
        \caption{\scriptsize ``autmobile'': CIFS-softplus}
    \end{subfigure}
    \begin{subfigure}{\SubFigWidth\linewidth}
        \centering
        \includegraphics[width=\SubImgWidth \linewidth]{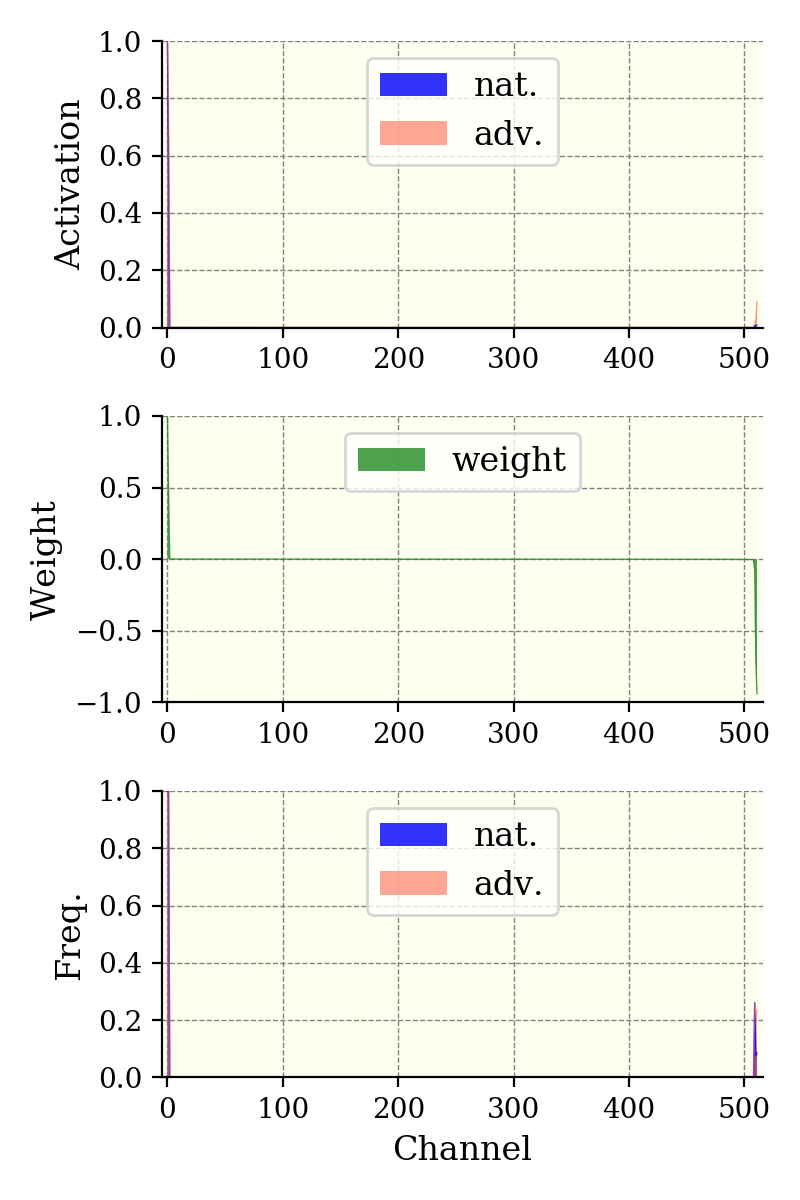}
        \vspace{-2em}
        \caption{\scriptsize ``autmobile'', CIFS-softmax}
    \end{subfigure}
    
    \begin{subfigure}{\SubFigWidth\linewidth}
        \centering
        \includegraphics[width=\SubImgWidth \linewidth]{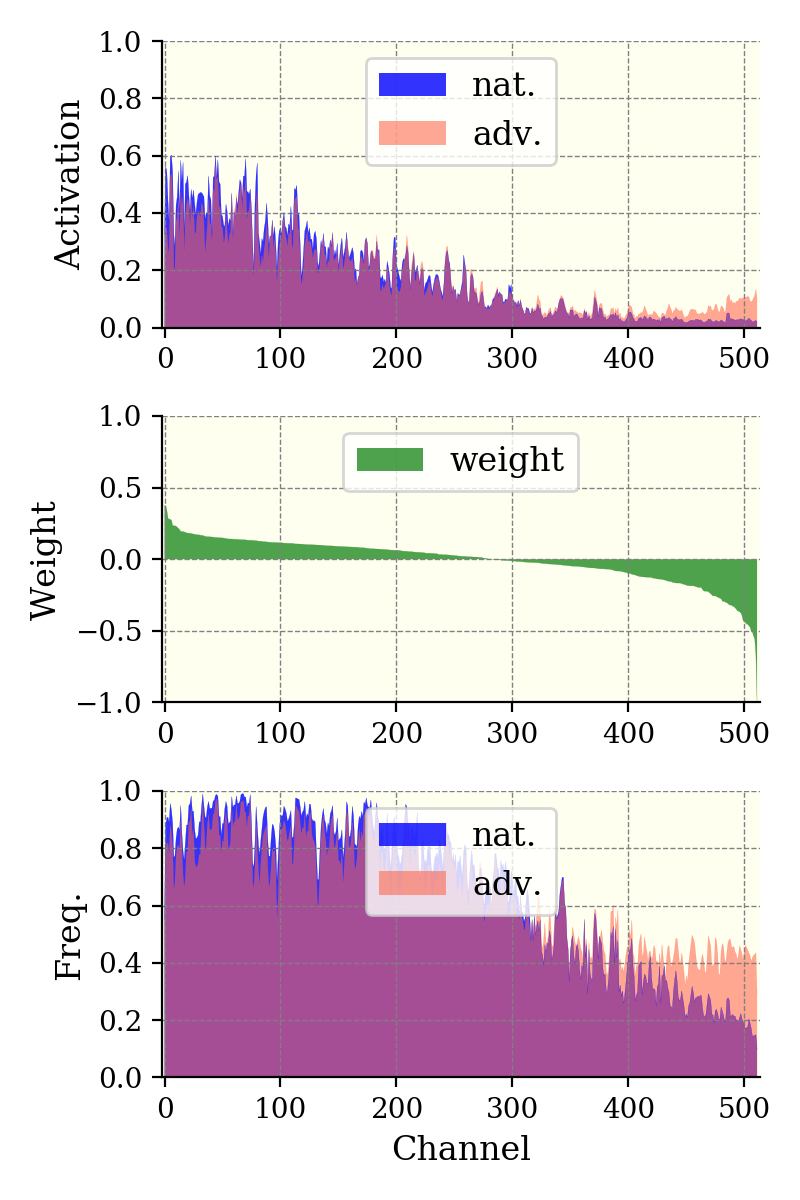}
        \vspace{-2em}
        \caption{\scriptsize ``frog'': non-CIFS}
    \end{subfigure}
    \begin{subfigure}{\SubFigWidth\linewidth}
        \centering
        \includegraphics[width=\SubImgWidth \linewidth]{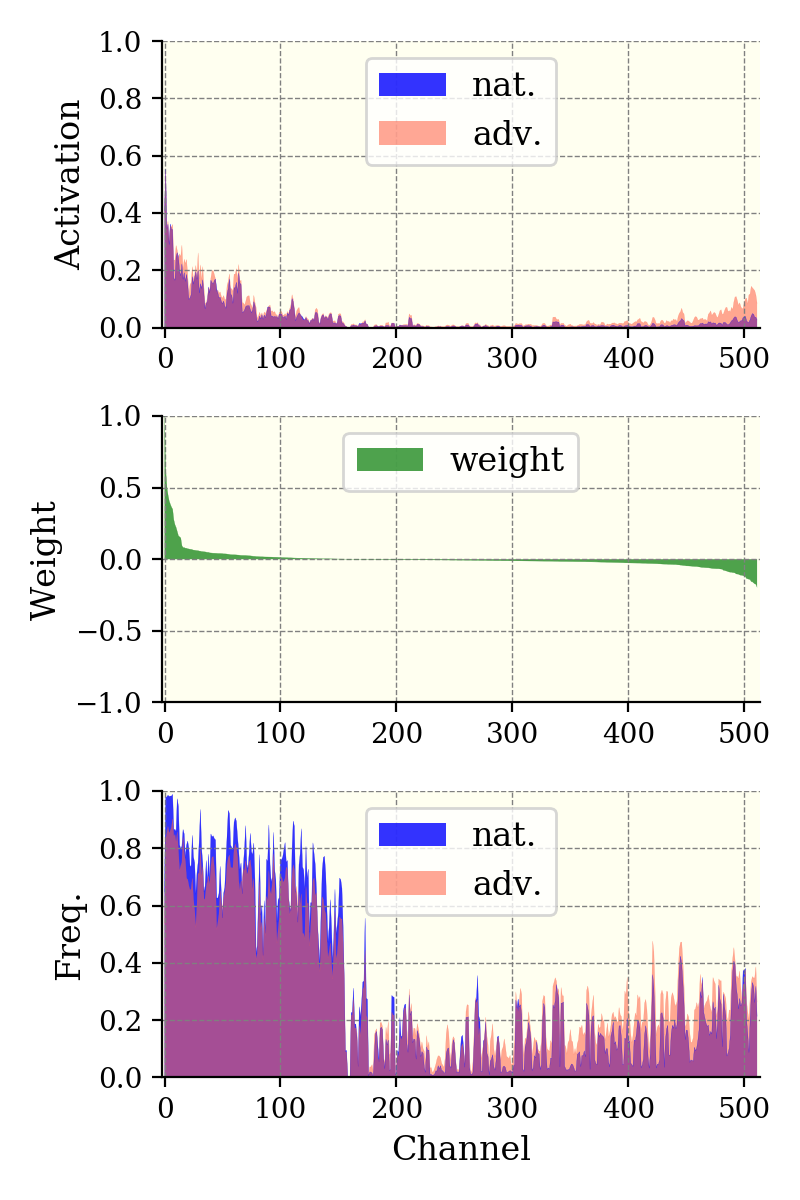}
        \vspace{-2em}
        \caption{\scriptsize ``frog'': CIFS-sigmoid}
    \end{subfigure}
    \begin{subfigure}{\SubFigWidth\linewidth}
        \centering
        \includegraphics[width=\SubImgWidth \linewidth]{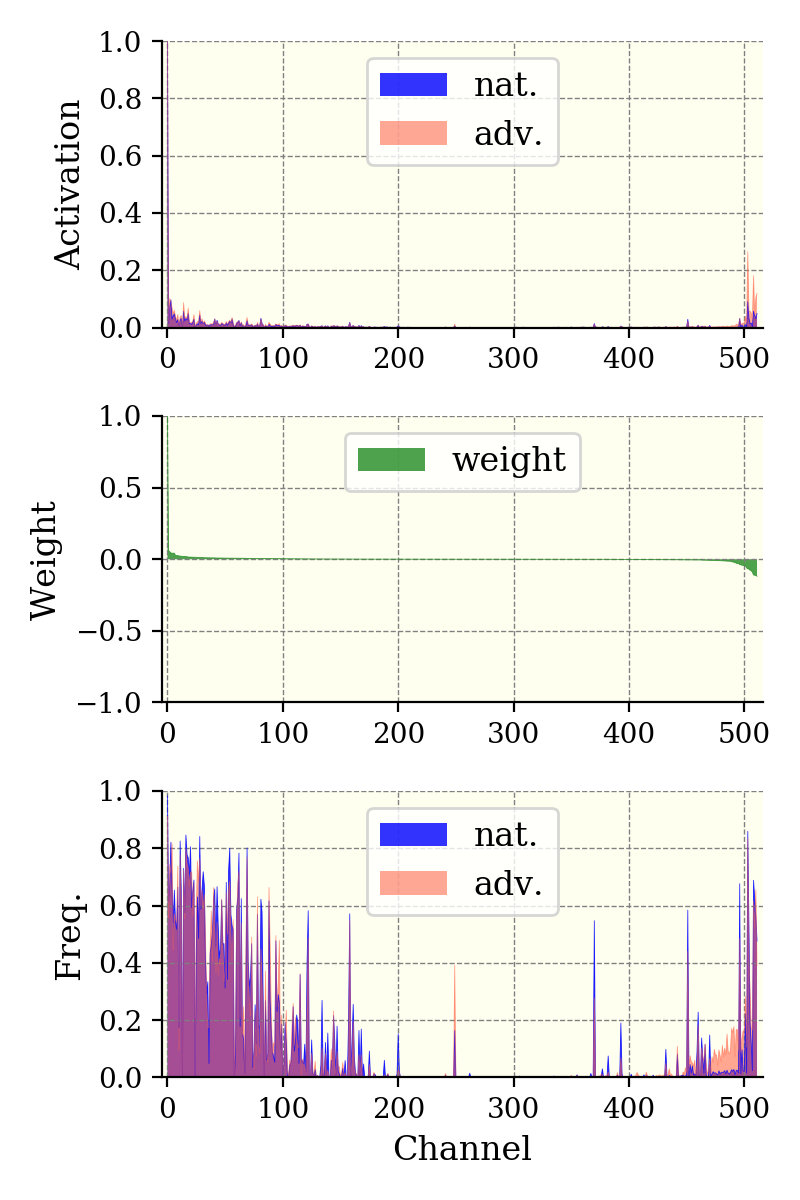}
        \vspace{-2em}
        \caption{\scriptsize ``frog'': CIFS-softplus}
    \end{subfigure}
    \begin{subfigure}{\SubFigWidth\linewidth}
        \centering
        \includegraphics[width=\SubImgWidth \linewidth]{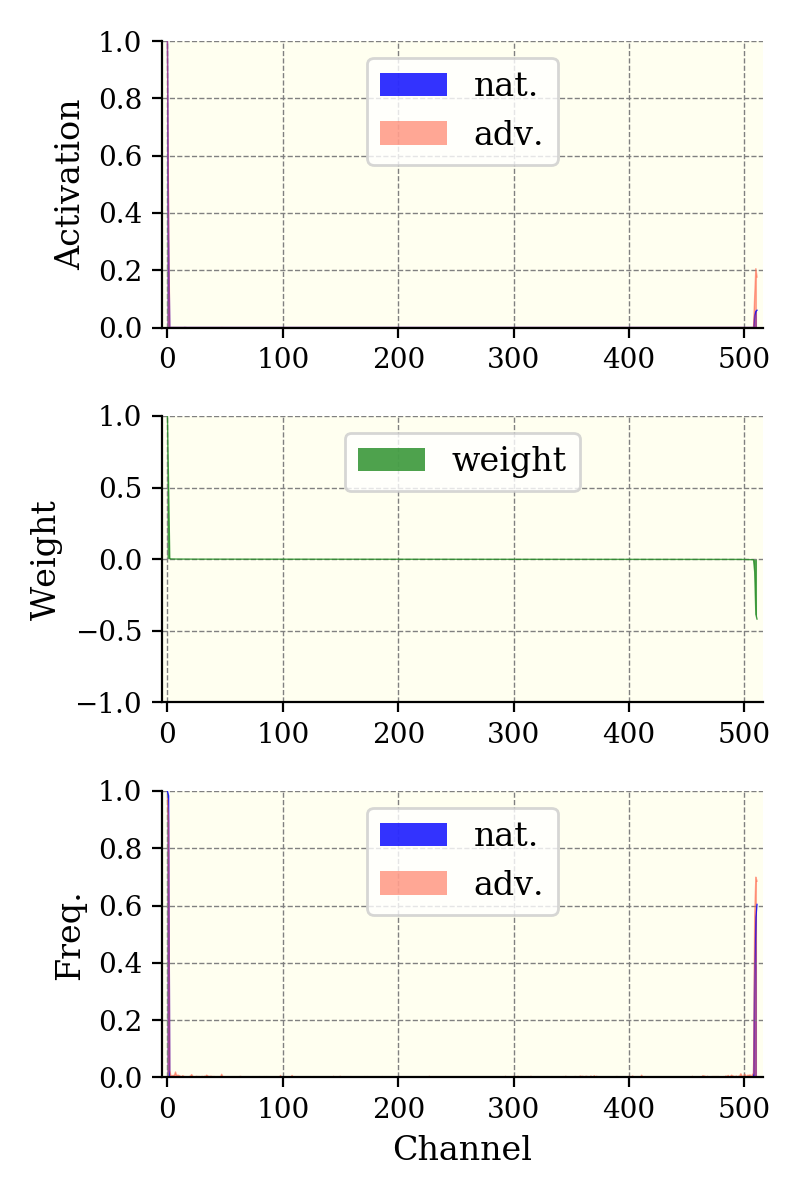}
        \vspace{-2em}
        \caption{\scriptsize ``frog'', CIFS-softmax}
    \end{subfigure}
    
    \caption{The magnitudes of channel-wise activations (top) at the penultimate layer, their activated frequency (bottom), and the weights of the last linear layer (middle) \textit{vs.} channel indices. The robust accuracies against PGD-20 (on the whole dataset) are 46.64\% for non-CIFS, 49.87\% for the CIFS-sigmoid, 50.38\% for the CIFS-softplus, and 51.23\% for the CIFS-softmax respectively}
    \label{fig:apx-hyp}
    \vspace{-1em}
\end{figure*}

\section{Robustness Evaluation on CIFAR10} \label{apdx:cifar10}

\subsection{Robustness Enhancement of CIFS under AT} \label{apdx:cifar10-at}
\textbf{Training and Evaluation details}: On the CIFAR10 dataset, we train ResNet-18 and WRN-28-10 models with PGD-10 adversarial examples ($\epsilon=\nicefrac{8}{255}$, step size $\nicefrac{\epsilon}{4}$ with random initialization). The $\beta$ in CIFS is set to be $2$. For the ResNet-18 and its CIFS-modified version, we train models for $120$ epochs with the SGD optimizer (momentum $0.9$ and weight decay $0.0002$). The learning rate starts from $0.1$ and is multiplied with $0.1$ at epoch $75$ and epoch $90$. For the WRN-28-10, we train model for $110$ epochs with weight decay $0.0005$. 

In Section \textcolor{Blue}{4.1}, we evaluate the robustness of CNNs against four white-box attacks with a perturbation budget $\epsilon=\nicefrac{8}{255}$ in $l_{\infty}$ norm --- FGSM, PGD-20 (step size $\nicefrac{\epsilon}{10}$), C\&W (optimized by PGD for 30 steps with a step size $\nicefrac{\epsilon}{10}$) and PGD-100 (step size $\nicefrac{\epsilon}{10}$).

\textbf{Robustness Evaluation with AutoAttack}: 
Here, we also report the robust accuracies of defense methods against AutoAttack \cite{croce_reliable_2020}, which consists of both white-box and black-box attacks. AutoAttack regards models to be robust at a certain data point only if the models correctly classify all types of adversarial examples generated by AutoAttack of that data point. We consider the AutoAttack including one strong white-box attack (Auto-PGD \cite{croce_reliable_2020}) and one black-box attack (Square-Attack \cite{andriushchenko2020square}). Since the Square Attack requires many queries, we sample \num[group-separator={,}]{2000} images (200 per class) from the CIFAR10 for evaluation. The attack parameters are set according to the officially released AutoAttack\footnote{\url{https://github.com/fra31/auto-attack}}. From Table \ref{tab:apdx-aa-cifar}, we observe that CIFS enjoys better robustness against AutoAttack in comparison to the vanilla ResNet-18 model and its CAS-modified version. 

\begin{table}[t!]
    \centering
    \caption{Robustness comparison of defense methods on CIFAR10. We report the last-epoch robust accuracies (\%) against AutoAttack.}
    \label{tab:apdx-aa-cifar}
    \vspace{1em}
    \scalebox{.8}{
    \begin{tabular}{cccc}
    \toprule
    \tbf{\tit{ResNet-18}} & Vanilla & CAS & CIFS  \\
    ResNet-18 & 44.00 & 42.70 & \tbf{46.20} \\
    WRN-28-10 & 47.20 & 46.55 & \tbf{49.75} \\
    \bottomrule
    \end{tabular}
    }
\end{table}{}

\textbf{Best-epoch robustness during training}: 
Due to the susceptibility of overtrained models to overfitting \cite{rice_overfitting_2020}, it seems reasonable to compare the results at the end of the training (and not for the  best epochs) \cite{madry_towards_2019, zhang_attacks_2020, rice_overfitting_2020}. In Section \textcolor{Blue}{4.1}, we report the robust accuracies of ResNet-18 and WRN-28-10 models at the last epochs. Here, we also provide the results at the \tbf{best} epochs for reference. 

From Table \ref{tab:apdx-cifar-best}, we see that, for the ResNet-18 architecture, the CIFS-modified model results in the similar best-epoch robustness (PGD-100) to that of the vanilla ResNet-18. For the WRN-28-10, the vanilla model has the better best-epoch robustness compared to the CIFS-modified version. This may be due to the fact that CIFS suppresses redundant channels and reduces the model capacity. 

By comparing results in Table \ref{tab:apdx-cifar-best} with those in Table \textcolor{Blue}{1}, we observe that CIFS indeed ameliorates the overfitting of AT. Specifically, the best-epoch robust accuracy of the vanilla WRN-28-10 (resp. ResNet-18) against PGD-100 attack is 54.17\% (resp.49.47\%), but the last-epoch accuracy drops to 47.08\% (resp. 44.72\%). In contrast, for the CIFS-modified versions, the last-epoch robust accuracies against PGD-100 attack are maintained around the best-epoch ones (for WRN-28-10, from 52.03\% to 51.51\%; for ResNet-18, from 49.76\% to 48.74\%). 

\begin{table}[h!]
    \centering
    \caption{Robustness comparison of defense methods on CIFAR10. We report the best robust accuracies (\%) during training. For each model, the results of the strongest attacks are marked with an underline.}
    \label{tab:apdx-cifar-best}
    \vspace{1em}
    
    \scalebox{.8}{
    \begin{tabular}{cccccc}
    \toprule
    \tbf{\tit{ResNet-18}} & Natural & FGSM & PGD-20 & C\&W & PGD-100 \\
    \hline
    Vanilla & 83.63 & 56.73 & 50.64 & 49.51 & \underline{49.47} \\
    CAS & 85.66 & 56.25 & 47.69 & 46.52 & \underline{45.69} \\
    CIFS & 82.46 & \tbf{58.98} & \tbf{51.94} & \tbf{51.25} & \tbf{\underline{49.76}} \\
    \bottomrule
    \toprule
    \tbf{\tit{WRN-28-10}} & Natural & FGSM & PGD-20 & C\&W & PGD-100\\
    \hline
    Vanilla & 86.53 & \tbf{61.43} & \tbf{55.69} & \tbf{54.45} & \tbf{\underline{54.17}} \\
    CAS & 87.51 & 58.54 & 52.06 & 51.27 & \underline{50.69} \\
    CIFS & 84.67 & 61.03 & 54.09 & 53.76 & \underline{52.03} \\
    \bottomrule
    \end{tabular}
    }
\end{table}{}


\textbf{Robust accuracies for various values of $\beta_{\text{atk}}$:} In Section \textcolor{Blue}{4.1}, we evaluate the robustness of CIFS-modified models by using the adaptive loss in Equation (\textcolor{Blue}{3}). For each type of attack, we assign various values to $\beta_{\text{atk}}$ and report the worst robust accuracies. Here, for reference, we provide the defense results of the ResNet-18 model on CIFAR10 for different values of $\beta_{\text{atk}}$ that are used in Section \textcolor{Blue}{4.1}. The results in Table \textcolor{Blue}{1} (ResNet-18) are collected from Table \ref{tab:apdx-beta-attack}. 

\begin{table}[t!]
    \vspace{-1em}
    \centering
    \caption{Robust accuracies (\%) for values of $\beta_{\text{atk}}$ on CIFAR10. The value ``$\infty$'' means the attack only considers the second term in Equation (\textcolor{Blue}{3}). The value ``$\infty$-1'' (resp. ``$\infty$-2'') means the attacker completely focuses on the first (resp. second) CIFS-modifed layer. The bracketed numbers are those reported in Table \textcolor{Blue}{1} (ResNet-18).}
    \vspace{.5em}
    \scalebox{.8}{
    \begin{tabular}{ccccccc}
    \toprule
    \tbf{\tit{ResNet-18}} & $\beta_{\text{atk}}$ & Natural &  FGSM & PGD-20 & C\&W & PGD-100 \\
    \midrule
    Vanilla & - & [84.56] & [55.11] & [46.62] & [45.95] & [\underline{44.72}] \\
    \hline
    CAS     & 0& [86.73] & 83.17 & 88.45 & 88.52 & 88.24 \\
    ~ & 0.1 & - & 58.61 & 61.36 & 85.51 & 62.40\\
    ~ & 1 & - & 56.36 & 52.86 & 62.34 & 56.02\\
    ~ & 2 & - & 56.06 & 49.76 & 54.94 & 50.62\\
    ~ & 10 & - & 56.03 & 47.47 & 49.35 & 47.70\\
    ~ & 100 & -  & 56.02 & 47.04 & 48.36 & 46.74\\
    ~ & $\infty$ & - & 56.02 & 47.06 & 48.31 & 46.55\\
    ~ & $\infty$-1 & - & [55.99] & [45.29] & [44.18] & [\underline{43.22}] \\
    ~ & $\infty$-2 & - & 82.68 & 87.87 & 87.79 & 87.72\\
    \hline
    CIFS & 0 & [83.86] & 60.58 & 52.64 & 51.32 & 49.94 \\
    ~ & 0.1 & - & [\tbf{58.86}] & 51.40 & 50.88 & 49.42\\
    ~ & 1 & - & 59.20 & 51.28 & [\tbf{50.16}] & 48.74\\
    ~ & 2 & - & 59.24 & [\tbf{51.23}] & 50.28 & 48.79 \\
    ~ & 10 & - & 59.35 & 51.27 & 50.70 & [\underline{\tbf{48.70}}] \\
    ~ & 100 & - & 59.38 & 51.41 & 51.04 & 48.80\\
    ~ & $\infty$ & - & 59.43 & 51.45 & 51.08 & 48.82\\
    ~ & $\infty$-1 & - & 61.06 & 54.96 & 53.83 & 52.82\\
    ~ & $\infty$-2 & - & 60.03 & 52.30 & 50.92 & 50.03\\
    
    \bottomrule
    \end{tabular}
    }
    \label{tab:apdx-beta-attack}
    \vspace{-1em}
\end{table}{}

\subsection{Robustness Enhancement under TRADES} \label{apdx:cifar10-trades}
To improve the robustness of CNNs, various training-based strategies have been proposed, including vanilla adversarial training (AT) \cite{madry_towards_2019}, friendly-adversarial training (FAT) \cite{zhang_attacks_2020}, and TRADES \cite{zhang_theoretically_2019}. In Section \textcolor{Blue}{4.1}, we show that CIFS can further enhance the robustness of CNNs under the vanilla AT and FAT. Here, we conduct more experiments to check whether TRADES is also suitable for CIFS.

\begin{table}[h!]
    \centering
    \caption{Robustness comparison of vanilla CNNs and their CIFS-modified version under various AT-based strategies. We report the robust accuracies (\%) on various types of adversarial data.}
    \label{tab:apdx-trades}
    \vspace{1em}
    \scalebox{.8}{
    \begin{tabular}{ccccc}
    \toprule
    \tbf{\tit{ResNet-18}} & Natural & FGSM & PGD-20 & PGD-100 \\
    \hline
    Vanilla-AT & 84.56 & 55.11 & 46.62 & \underline{44.72} \\
    Vanilla-TRADES & 83.96 & 57.09 & 50.27 & \underline{{48.83}}  \\
    Vanilla-FAT & 87.16 & 56.43 & 47.64  & \underline{45.35} \\
    \hline
    CIFS-AT     & 83.86 & {58.86} & {51.23} & \underline{{48.74}} \\
    CIFS-TRADES     & 85.20 & 54.76 & 46.13 & \underline{43.65}  \\
    CIFS-FAT & 86.35 & \tbf{59.47} & \tbf{51.68} & \underline{\tbf{49.52}}\\
    \bottomrule
    \end{tabular}
    }
\end{table}{}

From Table \ref{tab:apdx-trades}, we observe that, for the vanilla ResNet-18 model, TRADES effectively robustifies the network and outperforms its counterparts by a large margin (e.g., 48.83\% of TRADES \tit{vs.} 44.72\% of AT against PGD-100 attack). However, for the CIFS-modified models, TRADES performs worse than AT and FAT. In general, CIFS-modification in combination with the FAT training strategy achieves the best robustness against various attacks.

\section{Robustness Evaluation on SVHN} \label{apdx:svhn}
\textbf{Training and Evaluation details}: On the SVHN dataset, we train the ResNet-18 model and its CIFS-modified version with PGD-10 adversarial examples ($\epsilon=\nicefrac{8}{255}$, step size $\nicefrac{\epsilon}{4}$ with random initialization). We train models for $120$ epochs with the SGD optimizer (momentum $0.9$ and weight decay $0.0005$). The learning rate starts from $0.01$ and is multiplied with $0.1$ at epoch $75$ and epoch $90$.

In Section \textcolor{Blue}{4.1}, we evaluate the robustness of CNNs against four white-box attacks with a perturbation budget $\epsilon=\nicefrac{8}{255}$ in $l_{\infty}$ norm --- FGSM, PGD-20 (step size $\nicefrac{\epsilon}{10}$), C\&W (optimized by PGD for 30 steps with a step size $\nicefrac{\epsilon}{10}$) and PGD-100 (step size $\nicefrac{\epsilon}{10}$).

\textbf{Robustness Evaluation with AutoAttack}: 
Here, we also report the robust accuracies of defense methods against AutoAttack on SVHN (Table \ref{tab:apdx-aa-svhn}). The evaluation settings of AutoAttack follows those in Appendix \ref{apdx:cifar10-at}.

\begin{table}[h!]
    \centering
    \caption{Robustness comparison of defense methods on SVHN. We report the robust accuracies (\%) at the last epochs.}
    \label{tab:apdx-aa-svhn}
    \vspace{1em}
    \scalebox{.8}{
    \begin{tabular}{cccc}
    \toprule
    \tbf{\tit{ResNet-18}} & Vanilla & CAS & CIFS  \\
    AutoAttack & 40.60 & 39.30 & \tbf{42.10} \\
    \bottomrule
    \end{tabular}
    }
\end{table}{}

\tbf{Best-epoch robustness during training:} In Section \textcolor{Blue}{4.1}, we report the robust accuracies of ResNet-18 models at the last epochs during training. Here, we report the best-epoch robustness for reference (Table \ref{tab:apdx-svhn-best}). We see that CIFS modified version enjoys the better best-epoch robustness in comparison to the vanilla ResNet-18 model.
\begin{table}[h!]
    \centering
    \caption{Robustness comparison of defense methods on SVHN. We report the best robust accuracies (\%) during training. For each model, the results of the strongest attack are marked with an underline.}
    \label{tab:apdx-svhn-best}
    \vspace{1em}
    \scalebox{.8}{
    \begin{tabular}{cccccc}
    \toprule
    \tbf{\tit{ResNet-18}} & Natural & FGSM & PGD-20 & C\&W & PGD-100 \\
    \hline
    Vanilla & 93.88 & 66.02 & 51.71 & 48.87 & \underline{47.59} \\
    CAS & 93.90 & 65.53 & 50.52 & 48.39 & \underline{46.39} \\
    CIFS & 93.27 & \tbf{67.36} & \tbf{52.67} & \tbf{50.20} & \tbf{\underline{48.36}} \\
    \bottomrule
    \end{tabular}
    }
\end{table}{}

\section{More Results on FMNIST} \label{apdx:fmnist}
\textbf{Training and Evaluation details:} On the FMNIST dataset, we train ResNet-10 with PGD-20 adversarial examples ($\epsilon=0.3$, step size $0.02$ with random initialization). The $\beta$ in CIFS is set to be $2$. We train models for $120$ epochs with the SGD optimizer (momentum $0.9$ and weight decay $0.0002$). The learning rate starts with $0.1$ and is multiplied with $0.1$ at epochs $45$, $75$ and $90$.

We evaluate the robustness of the ResNet-10 models against FGSM, PGD-20, and PGD-100 white-box attacks. The perturbation is bounded by $\epsilon=0.3$ in $l_{\infty}$ norm. The step size of PGD-20 is set to be $0.01$, and that of PGD-100 is set to be $0.02$. Here, we report both the last-epoch robust accuracies and the best-epoch robust accuracies in Table \ref{tab:apdx-fmnist-robustness}.

\begin{table}[h!]
    \centering
    \caption{Robustness comparison of defense methods on FMNIST. For each model, the robust accuracies (\%) of the strongest attack are remarked with an underline. For each type of attack, the best defense results are highlighted in bold. Comparing the defense rates of the strongest attacks, we observe that CIFS outperforms other defenses by a large margin.}
    \label{tab:apdx-fmnist-robustness}
    \vspace{1em}
    
    \scalebox{.8}{
    \begin{tabular}{cccccc}
    \toprule
    \tbf{\tit{Last}} & Natural & FGSM & PGD-40 & PGD-100 \\
    \hline
    Vanilla & 85.19 & {80.52} & 66.47 & \underline{60.99} \\
    CAS & 86.59 & \tbf{82.45} & 65.58 & \underline{59.51} \\
    CIFS & 83.35     & 77.48 & \tbf{66.59} & \underline{\tbf{65.50}} \\
    \bottomrule
    \toprule
    \tbf{\tit{Best}} & Natural & FGSM & PGD-40 & PGD-100 \\
    \hline
    Vanilla & 85.19 & {81.21} & 67.63 & \underline{63.36} \\
    CAS & 86.63 & \tbf{83.59} & 68.73 & \underline{62.65} \\
    CIFS & 83.32     & 78.55 & \tbf{69.05} & \underline{\tbf{67.21}} \\
    \bottomrule
    \end{tabular}
    }
\end{table}

\section{More Results on Ablation Study} \label{apdx:ablation}

\subsection{Effects of $\beta$ in CIFS:} \label{apdx:ablation-beta}

Here, we train CIFS-modified ResNet-18 models on CIFAR10 with various values of $\beta$ in Equation (\textcolor{Blue}{3}). The coefficient $\beta$ balances the accuracies of raw predictions and the final prediction. From Table \ref{tab:apdx-cifar-beta}, we observe that $\beta$ values that are too small or too large values lead to drops in the accuracies of natural data and adversarial data. On the one hand, if the value of $\beta$ is too small, the raw predictions made by CIFS are not reliable. Thus, the channels selected by CIFS may not be the truly relevant ones with respect to the ground-truth class. On the other hand, if the value of $\beta$ is too large, the optimization procedure mostly considers the raw predictions, the final prediction (output) becomes unreliable. When $\beta=2$, we achieve the best robustness against various types of attack.

\begin{table}[h!]
    \centering
    \caption{Robustness accuracies (\%) on CIFAR10 for CIFS with various values of $\beta$.}
    \label{tab:apdx-cifar-beta}
    \vspace{1em}
    
    \scalebox{.8}{
    \begin{tabular}{ccccc}
    \toprule
    \tbf{\tit{ResNet-18}} & Natural & FGSM & PGD-20 & PGD-100 \\
    \hline
    Vanilla & 84.56 & 55.11 & 46.62 & \underline{44.72} \\
    $\beta=0.1$ & 75.22 & 53.41 & 48.10 & \underline{46.28} \\
    $\beta=1$ & 82.34 &58.15 & 50.50 & \underline{48.35}\\
    $\beta=2$ & 83.86 & \tbf{58.86} & \tbf{51.23} & \underline{\tbf{48.74}} \\
    $\beta=10$ & 82.97 & 57.62 & 49.34 & \underline{47.10}\\
    $\beta=100$ & 75.41 & 52.90 & 45.00 & \underline{43.12}\\
    
    \bottomrule
    \end{tabular}
    }
\end{table}{}

\subsection{Effects of the top-$k$ feature assessment}

In general, $k$ should be larger than 1 but not too large. 

If we use the top-$1$, once adv. data fool probe nets, the channels relevant to true labels will be missed, and this will lead to wrong predictions (Table \textcolor{Blue}{5}, line top-1). Instead, we use top-$2$ for reliable channel selection. The efficacy is attributed to \tit{two} aspects: {Firstly}, the top-$2$ accuracies of adv. data are usually high (see Table \textcolor{Blue}{4}), thus channels relevant to top-$2$ logits \tit{include those relevant to the true class}. 
{Secondly}, \textcolor{black}{\textcolor{Blue}{Tian et al.} (\textcolor{Blue}{2021})}\footnote{\scriptsize Q. Tian, K, Kuang, F. Wu, Y. Wang, Intriguing class-wise properties of adversarial training. OpenReview. 2021} reports that CNNs' predictions of adv. data usually belong to the superclass that contains true labels. Classes (e.g., cat, dog) in the same superclass (e.g., animals) usually share similar semantic features. Thus, {most of the top-$2$ selected channels are useful} for predicting the true class. 

Although the top-$2$ selected channels may contain info about the other wrong class, the following layers (after CIFS) are capable of “purifying” features and make better predictions. This is verified by Table \textcolor{Blue}{5}, the results in the line top-2 (CIFS/CIFS 48.72\% vs. CIFS/Final 54.96\%) mean that around 6\% adv. data, which successfully fool probes, are still finally correctly classified. However, too large $k$ may degrade the relevance assessment due to too much noisy info (e.g., the effect of top-3 is worse than top-2 in Table \textcolor{Blue}{5}).

\subsection{Layers to be modified}

\tbf{Positions of CIFS modules}: Here, we try different combinations of the layers to be modified by CIFS. In CNNs, the features of deep layers are usually more characteristic in comparison to those in the shallower layers \cite{zeiler_visualizing_2014}, and each channel of the features captures a distinct view of the input. The predictions often depend only on the information of a few essential views of the inputs. CIFS improves adversarial robustness by adjusting channel-wise activations. Thus, we apply CIFS to the deeper layers instead of the shallower ones. Specifically, we modify the ResNet-18 by applying CIFS at the last (P1) and/or the second last (P2) residual blocks. The experimental results are reported in Table \ref{tab:positions}. We observe that simultaneously applying CIFS into P1 and P2 performs the best against various attacks. Intuitively, because the features can be progressively refined, applying CIFS at P1\&P2 better purifies the channels compared to applying it only at P1 or P2.

\begin{table}[h]
    \centering
    \caption{Robustness (\%) comparison of the positions where CIFS modules are placed.}
    \scalebox{.8}{
    \begin{tabular}{cccccc}
    \toprule
    \tbf{\tit{ResNet-18}} & Natural & FGSM & PGD-20 & PGD-100\\
    \hline
    Vanilla & 84.56 & 55.11 & 46.62 & \underline{44.72} \\
    P1 & 84.02 & 57.60 & 48.45 & \underline{45.95} \\
    P2 & 82.62 & 56.55 & 47.22 & \underline{44.81} \\
    P1-P2 & 83.86 & \tbf{58.86} & \tbf{51.23} & \underline{\tbf{48.74}} \\
    \bottomrule
    \end{tabular}
    }
    \label{tab:positions}
\end{table}{}

\subsection{Architecture of Probe Networks}

\tbf{Linear vs. Non-linear Probe}: For a certain layer modified by CIFS, the probe network in CIFS serves as the surrogate classifier of the subsequent layers in the backbone model. Thus, the probe networks should be powerful enough to make correct predictions based on the features of this layer. For the CIFS in the last residual block, we use a linear layer network as the probe, while for the CIFS in the second last residual block, we compare the cases of using a linear layer versus using a two-layer MLP network. From Table \ref{tab:linear}, we observe that the MLP-Linear combination shows a similar performance compared to the combination of two linear layers against adversarial attacks, but enjoys a clear advantage on the natural data ($83.86$\% vs. $81.52$\%). This is because the features in the second last residual block are not as characteristic as those in the last block and cannot be linearly separated. The MLP can thus classify the features better than a pure linear layer.

\begin{table}[h]
    \centering
    \caption{Robustness comparison (\%) of the probe architectures in CIFS modules (at P1-P2).}
    \vspace{.5em}
    \scalebox{.8}{
    \begin{tabular}{cccccc}
    \toprule
    \tbf{\tit{ResNet-18}} & Natural & FGSM & PGD-20 & PGD-100\\
    \hline
    Vanilla & 84.56 & 55.11 & 46.62 & \underline{44.72} \\
    Linear-Linear & 81.52 & 58.33 & \tbf{51.32} & \underline{\tbf{49.07}} \\
    MLP-Linear & 83.86 & \tbf{58.86} & 51.23 & \underline{48.74} \\
    \bottomrule
    \end{tabular}
    }
    \label{tab:linear}
\end{table}{}

%
%
%

\end{document}